%% file: main.tex
\documentclass[10pt,twocolumn,letterpaper]{article}

\usepackage[pagenumbers]{cvpr} %

\usepackage[accsupp]{axessibility}  %
\definecolor{cvprblue}{rgb}{0.21,0.49,0.74}
\usepackage[pagebackref,breaklinks,colorlinks,allcolors=cvprblue]{hyperref}

\usepackage{xcolor}
\usepackage{bbm}
\usepackage{booktabs}
\usepackage{graphicx}
\usepackage{multirow}
\usepackage{wrapfig}
\usepackage{tabularx}
\usepackage[normalem]{ulem}
\usepackage{svg}

\usepackage[capitalize]{cleveref}
\crefname{section}{Sec.}{Secs.}
\Crefname{section}{Section}{Sections}
\Crefname{table}{Table}{Tables}
\crefname{table}{Tab.}{Tabs.}

\newcommand{\name}{\emph{DepthCues}\xspace}

\title{\name: Evaluating Monocular Depth Perception in Large Vision Models}

\author{Duolikun Danier \quad Mehmet Ayg\"un \quad Changjian Li \quad Hakan Bilen \quad Oisin Mac Aodha  \\ [0.3em]
University of Edinburgh \\ [0.1em]
\small\url{https://danier97.github.io/depthcues}
}

\begin{document}
\maketitle

\begin{abstract}
Large-scale pre-trained vision models are becoming increasingly prevalent, offering expressive and generalizable visual representations that benefit various downstream tasks. Recent studies on the emergent properties of these models have revealed their high-level geometric understanding, in particular in the context of depth perception. However, it remains unclear how depth perception arises in these models without explicit depth supervision provided during pre-training. To investigate this, we examine whether the monocular depth cues, similar to those used by the human visual system, emerge in these models. We introduce a new benchmark, DepthCues, designed to evaluate depth cue understanding, and present findings across 20 diverse and representative pre-trained vision models. Our analysis shows that human-like depth cues emerge in more recent larger models. We also explore enhancing depth perception in large vision models by fine-tuning on DepthCues, and find that even without dense depth supervision, this improves depth estimation. To support further research, our benchmark and evaluation code will be made publicly available for studying depth perception in vision models.
\end{abstract}

\section{Introduction} \label{sec:intro}

To ensure safe, accurate, and robust interaction with the 3D world, it is important for computer vision models to understand geometric properties from their projections onto 2D images. Of particular significance is the perception of depth from single images, which has been found to benefit various downstream tasks including segmentation~\cite{laoviability}, object detection~\cite{huang2022monodtr}, and image and video generation~\cite{zhang2023adding, liang2024movideo}, among others. 
Monocular depth estimation models traditionally rely on dense depth supervision, but obtaining high-quality depth ground truth at scale is challenging, unlike the abundance of large 2D image-only datasets. 

Large-scale pre-trained vision models like  Stable Diffusion~\cite{rombach2022high} and DINOv2~\cite{oquab2024dinov}, trained solely on 2D images, have shown strong generalization when fine-tuned on various computer vision tasks, including  multi-view~\cite{izquierdo2025mvsanywhere,cao2024mvsformer} and monocular depth estimation~\cite{el2024probing, ge2024geobench, zhan2024physd}, even reaching human-level performance in ordering objects by depth~\cite{linsley20243d}. As a result, numerous recent monocular depth estimation approaches~\cite{gui2024depthfm, ke2024repurposing, depth_anything_v2} leverage these models' depth priors to achieve state-of-the-art performance. This naturally raises an important question: \textit{How does depth perception arise in these large vision models?}

\begin{figure}
    \centering
    \includegraphics[width=\linewidth]{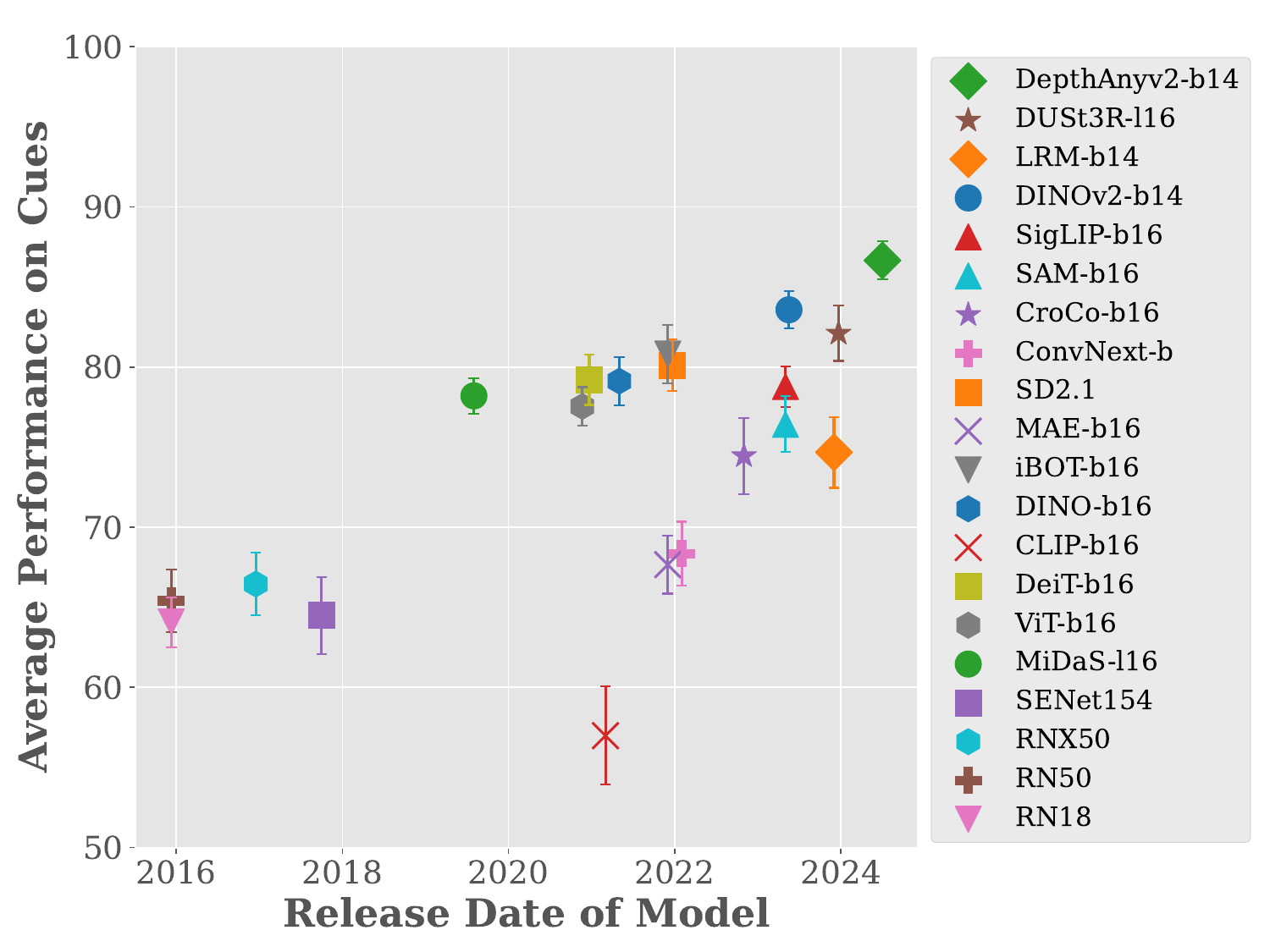}
    \vspace{-18pt}
    \caption{ 
        \textbf{Human-like monocular depth cues emerge in large vision models}. We present \name, a comprehensive benchmark suite designed to probe the understanding of human monocular depth cues in vision models. We analyse a diverse set of vision models and find that recent self-supervised and geometry estimation models demonstrate a notably stronger grasp of these cues, even in cases where models (\eg DINOv2) were not explicitly pre-trained on any depth-related tasks.
    }
    \label{fig:performance_vs_year}
    \vspace{-10pt}
\end{figure}

\begin{figure*}[ht]
    \includeinkscape[width=1.0\textwidth]{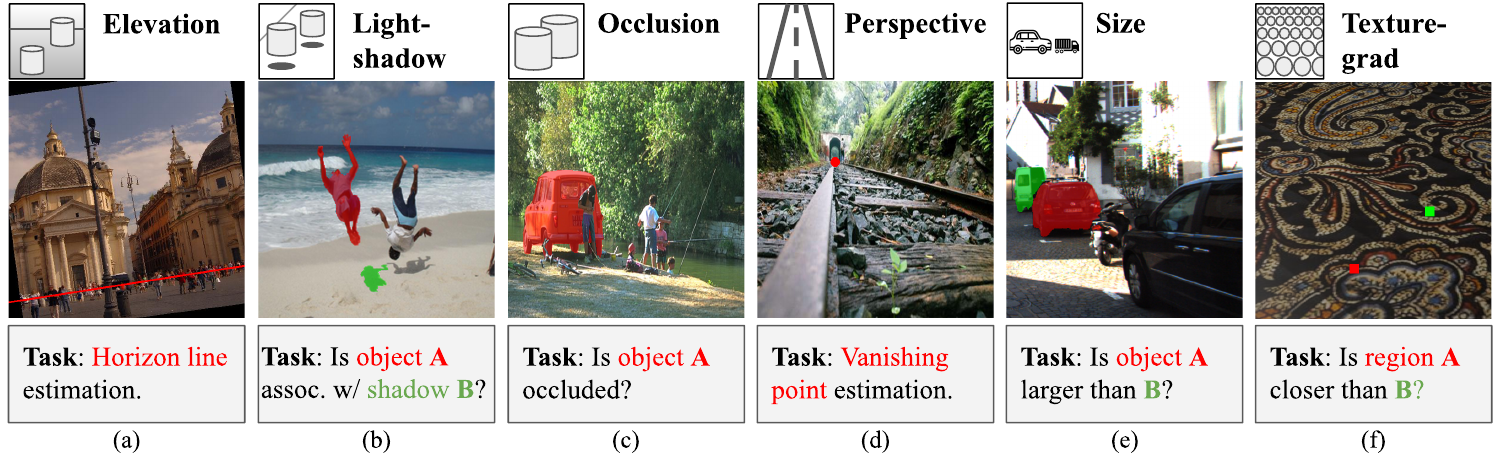}
    \vspace{-17pt}
    \caption{
        \textbf{Overview of \name.} Monocular depth cues, the associated tasks, and example instances from our proposed benchmark.
    }
    \label{fig:task_overview}
    \vspace{-8pt}
\end{figure*}

Studies on the human visual system~\cite{kalloniatis2011perception,bull2021intelligent} suggest that humans infer monocular depth using various visual cues such as elevation, light and shadow, occlusion, perspective, size, and texture gradient~\cite{bogdanova2016depth, lebreton2012perceptual} (see the top part of ~\cref{fig:task_overview}). 
Although these visual cues could emerge in large vision models during pre-training, it remains unclear to what extent this is true. 
Existing benchmarks~\cite{el2024probing,ge2024geobench,zhan2024physd, azad2024geometer, linsley20243d} focus on evaluating the depth estimation performance of large vision models but do not investigate their understanding of the underlying visual cues. 
To address this gap, \textbf{we evaluate how well the large vision models understand and utilize human-like monocular depth cues.} 
Our goal is to provide insights into current models' depth estimation behavior, enable a more systematic analysis of their strengths and limitations, and indicate areas for improvement.

We introduce a new benchmark named \textbf{\name} consisting of six depth related tasks
(see ~\cref{fig:task_overview}) where each task is designed to test a given models' ability to estimate a visual cue that is ubiquitous to human depth perception. Specifically, lightweight probing models (referred to as the \textit{probes} hereafter) placed on top of the frozen features of large vision models are trained and tested on the associated dataset for each task. 
The assumption is that the more effectively the probe can learn to solve a task using the features of a model, the better the model understands the task-specific cue. Under this protocol, we examine 20 vision models pre-trained at scale under various settings, including self-supervised learning (DINOv2~\cite{oquab2024dinov}), image generation (StableDiffusion)~\cite{rombach2022high}, vision-language supervision (CLIP~\cite{radford2021learning}), single-view reconstruction (LRM~\cite{hong2024lrm}), cross-view completion (CroCo~\cite{weinzaepfel2022croco}), and also a recent general purpose depth estimation model DepthAnythingv2~\cite{depth_anything_v2}. %

Our results reveal the following observations: First, the performance of models on \name is highly correlated with their depth estimation performance, validating the effectiveness of \name for investigating depth perception. 
Second, more recent self-supervised and geometry estimation models exhibit a stronger understanding of human monocular depth cues (see \cref{fig:performance_vs_year}), with DepthAnythingv2 ranking first, and DINOv2 and StableDiffusion ranking top-five. Third, none of the benchmarked models consistently excel on all six of the depth tasks. Finally, recent single-view pre-trained models are limited in their perception of texture gradient, while multi-view trained models are better at this low-level geometric cue. Additionally, we explore enhancing depth awareness of vision models by injecting depth cue priors, demonstrating that fine-tuning on our \name benchmark results in improved depth estimation, despite the significant sparsity of the supervision compared to dense annotations.
Our contributions are:
\begin{itemize}
    \item We develop and make available \name, a benchmark suite for evaluating the emergence of human monocular depth cues in large vision models. 
    \item We evaluate 20 diverse vision models across a range of pre-training settings, and analyze their relative strengths and weaknesses in capturing monocular depth cues.
    \item We show that human depth perception cues emerge from self-supervised pre-training on single images, with stronger alignment found in more recent, larger models.
    \item We conduct an exploratory study on enhancing the depth awareness of a model by fine-tuning on \name, showing that explicitly injecting human-like depth priors can improve depth perception.
\end{itemize}

\section{Related Work}

\noindent{\bf Monocular Depth Estimation.} Monocular depth estimation is an inherently ill-posed and challenging problem, yet deep learning-based methods have made significant strides in addressing it. 
Initial progress was driven by supervised methods~\cite{eigen2014depth,fu2018deep,bhat2021adabins} that were trained on various labeled datasets~\cite{Silberman_ECCV12,geiger2012we}. 
Later, self-supervised approaches advanced the field further by leveraging stereo~\cite{garg2016unsupervised,godard2017unsupervised} or temporal consistency~\cite{zhou2017unsupervised,gordon2019depth,godard2019digging}.
More recently, general purpose methods~\cite{ranftl2020towards,depth_anything_v1,depth_anything_v2,ke2024repurposing} leveraging large vision backbone models pre-trained with self-supervised~\cite{oquab2024dinov} or generative objectives~\cite{rombach2022high} and then fine-tuned on very large-scale supervised depth datasets have been shown to result in strong monocular depth estimation performance. 
However, it remains unclear how depth perception develops in the pre-trained models without explicit depth supervision during pre-training. In this paper, we try to shed light on the emergence of these monocular depth cues used by the human visual system in large vision models.

\noindent{\bf Depth Perception in Pre-trained Vision Models. } 
Several recent efforts have begun to analyze the 3D and depth perception abilities of pre-trained large vision models, with a focus on 3D awareness~\cite{el2024probing}, 3D scene understanding~\cite{man2024lexicon3d}, depth and height perception~\cite{azad2024geometer}, and 3D physical understanding~\cite{zhan2024physd}. 
Complementary to these efforts, several studies have attempted to enhance the capabilities of these models by integrating 3D and depth information~\cite{li24multitask,aygun2024enhancing,zhang2024condense,yue2024improving}. Unlike these works, we focus on monocular depth estimation, emphasising evaluating the emergence of monocular cues essential for human depth perception. Some of the cues we study are also explored in a concurrent work~\cite{zhan2024physd}, however, we focus on depth perception with diverse data and analyze a broader range of vision models.

Recent work~\cite{spencer2022deconstructing,ge2024geobench} has analyzed monocular depth estimation models, examining properties like the backbone architectures, loss functions, and training datasets. However, they do not explore whether or how these models actually utilize monocular cues for depth estimation. In our work, we focus on analyzing the existence of intrinsic depth cues without direct depth supervision across various large vision models. Closely related to our approach, \cite{theiner2023analyzing} investigates whether predictions of a monocular depth estimation model~\cite{ranftl2020towards} violate perceptual cues. However, a prediction that violates a perceptual cue does not necessarily indicate the absence of the cue in the model itself. Instead, we directly examine the presence of these cues across multiple models, including, but not limited to, monocular depth models using our newly introduced benchmark suite.

\noindent{\bf Comparing Humans to Neural Networks. } Human and animal vision systems have long inspired computational vision methods~\cite{hubel1968receptive,o1978hippocampus,hebb2005organization}. Many studies have sought to compare biological vision with modern neural network-based computer vision systems across tasks such as object recognition~\cite{rajalingham2018large}, shape-bias~\cite{geirhos2018imagenettrained}, multi-view consistency~\cite{o2023approaching, bonnen2024mochi}, visual perspective-taking~\cite{linsley20243d}, and perceptual judgments of similarity~\cite{fu2024dreamsim}. 
In our work, rather than measuring alignment between human and large-scale vision systems, we focus on the emergence of monocular cues that are well-established in the human vision science community~\cite{kalloniatis2011perception,bull2021intelligent,bogdanova2016depth,lebreton2012perceptual}, \ie the cues that are crucial for monocular depth estimation in human vision. 
Recent work has incorporated additional \emph{high}-level human vision-inspired cues, specifically semantics, object size, and spatial relationships, in an attempt to improve automated monocular depth estimation~\cite{auty2022monocular,auty2024language}. These cues, extracted using pre-trained semantic segmentation and language embedding models, are shown to enhance depth prediction accuracy. 
We hope that our results and analyses will inspire similar approaches and foster new directions.

\begin{table}[t]
\centering
\resizebox{1.0\linewidth}{!}{
\begin{tabular}{l l r r r r}
\toprule
Depth Cue           & Source         & \#Images & \#Train & \#Val & \#Test \\ \midrule
Elevation      & HLW~\cite{workman2016hlw}            &    18,390      &     15,761     &   828     & 1,801  \\
Light-shadow & SOBA~\cite{Wang_2020_CVPR}           &    798      &    4,716      &   862     &   860     \\
Occlusion    & COCOA~\cite{zhu2017semantic}          &   5,573      &   24,402       &   12,906     &    14,862    \\
Perspective       & NaturalScene~\cite{zhou2017detecting}          &  2,275        &    2,000      &   125     &   150     \\
Size      & KITTI~\cite{geiger2012we}, SUN~\cite{song2015sun} &  2,717      & 1,986         &   330     &    712    \\
Texture-grad & DTD~\cite{cimpoi14describing}            &    6,000      &    4,000      &   1,000     &     1,000   \\
 \bottomrule
\end{tabular}
}
\vspace{-5pt}
\caption{\textbf{Summary of the datasets in \name benchmark}. We report total image and data statistics across all splits.}
\label{tab:summary_cue_task_dataset}
\vspace{-10pt}
\end{table}

\section{\name: Benchmark and Dataset} 

We introduce \textbf{\name}, a new benchmark to investigate how large-scale pre-trained image models perceive depth by assessing their understanding of a set of monocular depth cues used by humans to perceive depth: elevation, light and shadow, occlusion, perspective, size, and texture gradient~\cite{kalloniatis2011perception,bull2021intelligent}. Our benchmark comprises a suite of six tasks with constituent datasets specifically designed to capture the essence of each individual monocular depth cue. For each task, \name provides training and test sets with an evaluation protocol for image models, along with a performance metric to assess how well a given model understands each depth cue. \cref{fig:task_overview} illustrates the task for each cue, and shows a representative instance from the associated dataset. We summarize the statistics of our datasets in \cref{tab:summary_cue_task_dataset}. 
Next, we describe the construction of each task and the data used. In each case, we adapt existing datasets that have been designed for bespoke vision tasks. We use the original train, validation, and test splits of data sources and keep the classes balanced for binary tasks unless specified. 
Additional details about dataset construction are provided in \cref{supp:dataset}.

\subsection{Elevation} \label{sec:benchmark_elevation}
Objects on the ground plane are farther away as they get closer to the horizon.

\noindent{\bf Task.} 
The elevation cue relies on perceiving the horizon line, which is defined as the projection of all the points where light rays from the observer's viewpoint are tangent to Earth's surface. 
Therefore, we adopt the task of horizon line estimation~\cite{workman2016hlw, kluger2020temporally}. 
Note, the horizon line is not always visible, and is different from the skyline, which separates the sky and the entities on the ground. 
The task is posed as a regression problem where the goal is to estimate the target line, \ie its slope and y-intercept, in the image.

\noindent {\bf Dataset.} We adopt the HLW dataset~\cite{workman2016hlw}, which provides pairs of images and annotations of the horizon lines.
To increase the slope variation, we augment the images by rotating them (and the line annotations) randomly between $-30^{\circ}$ and $30^{\circ}$. An example from the dataset is shown in \cref{fig:task_overview}~(a).

\subsection{Light and Shadow}\label{sec:benchmark_lightshadow}
The way shadow is cast provides useful cue for orientation and depth of objects. For convenience, we refer to this cue as light-shadow.

\noindent{\bf Task.} To capture the understanding of this cue, we propose a task of associating shadows with objects.

\noindent{\bf Dataset.} The SOBA~\cite{Wang_2020_CVPR} dataset contains images with segmentation masks of objects and their associated shadows. From those, we create a dataset containing tuples of $(I, M_A, M_B, y)$, where $M_A$ is the mask of the object $A$ in image $I$, and $M_B$ is the mask of a shadow $B$ in the image. The label $y$ is 1 if $B$ is associated with $A$, and 0 otherwise. An example is depicted in \cref{fig:task_overview}~(b).

\subsection{Occlusion}\label{sec:benchmark_occlusion}
An object that partially blocks another, from the perspective of the viewer, is perceived as closer. 

\noindent{\bf Task.} To test whether models understand occlusion, we propose the task of identifying whether an object is occluded or not by another object.

\noindent{\bf Dataset.} We use the COCOA~\cite{zhu2017semantic} dataset, providing manually annotated segmentation masks for visible and invisible object parts with occlusion rates. From the original annotations, we construct a dataset that contains triplets of image $I$, the segmentation mask $M_A$ of the visible part of the object $A$, and the corresponding label $y$, which is 1 if the object is occluded, and 0 otherwise. In the example in \cref{fig:task_overview}~(c), the car highlighted with a red mask is occluded.

\subsection{Perspective}\label{sec:benchmark_perspective}
Parallel lines converging at infinity offer a strong cue for perceiving depth.

\noindent{\bf Task.} This cue is closely tied to vanishing point~\cite{liu2021vapid}, which is defined as the point on the image plane where 2D perspective projections of 3D parallel lines intersect. Therefore, we use the task of vanishing point estimation to investigate the understanding of perspective. While there exist different versions of the task that define different numbers of vanishing points~\cite{tong2022transformer, zhai2016detecting}, we adopt a simplified version that focuses on a single dominant vanishing point. 
Given an image $I$, the objective is to estimate the $\mathbf{p}=(x,y)$ coordinates of the vanishing point in the image coordinate system. 

\noindent{\bf Dataset.} The NaturalScene dataset~\cite{zhou2017detecting} 
contains images from diverse outdoor scenes with one manually annotated dominant vanishing point $\mathbf{p}$ in each. We use the original dataset and create validation and test sets by dividing the original validation set. An example is shown in ~\cref{fig:task_overview}~(d) with the vanishing point indicated by the red dot.

\subsection{Size} \label{sec:benchmark_size}
People infer depth based on known object sizes (in 3D) and their apparent retinal sizes (\eg an object farther away appears smaller). This cue is particularly effective when comparing objects of similar expected sizes. 

\noindent{\bf Task.} Using this cue requires prior knowledge of object sizes. Thus we propose a binary classification task: determining if object $A$ is larger than object $B$ in 3D. %

\noindent{\bf Dataset.} We use 3D object detection datasets KITTI~\cite{geiger2012we} and SUN-RGBD~\cite{song2015sun} which provide 3D bounding boxes of objects in outdoor and indoor scenes respectively.
Given an image $I$, we sample two objects, $A$ and $B$, and compare their sizes using the volumes of their 3D bounding boxes. Object masks $M_A$ and $M_B$ in image space are estimated using SAM~\cite{kirillov2023segment} with bounding box queries and additional filtering to ensure segmentation quality. The final dataset contains tuples $(I, M_A, M_B, y)$, where the label $y=1$ if object $A$ is larger, and $0$ otherwise. We filtered pairs where the volume difference was under a threshold to reduce errors caused by imprecise labeling. See \cref{fig:task_overview}~(e) for an example.

\subsection{Texture Gradient} \label{sec:benchmark_texturegrad}
When a region on a surface is further away from the viewer, its texture appears denser and more compressed, with a higher spatial frequency, compared to the same texture viewed nearby. We denote this cue as texture-grad.

\noindent{\bf Task.} This cue is especially helpful on textured surfaces. To assess the existence of this cue, we use the task of depth ordering of regions on a textured plane, where the model must infer depth based solely on texture.

\noindent{\bf Dataset.} To ensure that only the texture gradient cue is used for perceiving depth order, we created a synthetic dataset using Blender. First, we initialized a 3D plane, and applied textures sampled from the DTD dataset~\cite{cimpoi14describing} which contains various real-world, as well as synthetic, textures. Then, fixing both the viewing direction of the camera (to always point at the plane center) and its distance to the plane center, we randomly set the camera elevation angle between $30^{\circ}$ and $60^{\circ}$, and project the textured plane on the camera imaging plane to obtain an image with its depth map. 

From each image $I$, we randomly sample two equal-sized regions $A$  and $B$ (with their masks $M_A, M_B$), then set label $y$ to 1 if $A$ is closer than $B$ and $0$ otherwise. To avoid the region higher up in the image always being farther away, we apply random horizontal and vertical flips, so that the task cannot be solved solely from locations of the regions.
An example image is shown in \cref{fig:task_overview}~(f).

\section{Evaluation Protocol}

Following previous work~\cite{el2024probing,zhan2024physd}, we use probing of visual features as our evaluation protocol. Below, we explain how we obtain task-specific input features from visual models, how we probe these features, and how we measure success. 
An illustrative figure is included in \cref{supp:implementation_probing}.

\subsection{Task Specific Visual Features}

Given a pre-trained vision model $\phi(\cdot)$, we assess its understanding of a cue by training a lightweight probe $g_\theta$ with parameters $\theta$ on the target task, using the features extracted by $\phi$. We obtain features $\mathbf{f}$ for each task as described below. 

\noindent\textbf{Occlusion.} We extract a feature map $\phi (I)$ from the image $I$ and up-sample it to match the spatial size of the object mask $M_A$. We mask the feature map and apply spatial pooling:

\begin{equation}
    \mathbf{f} = \frac{\sum_{h,w} M_A\odot \mathrm{up}(\phi(I))}{\sum_{h,w} M_A}, \label{eqn:feat_mask1}
\end{equation}
where $h,w$ index the height and width of $I$, $\odot$ is element-wise multiplication, and $\mathrm{up}(\cdot)$ denotes bilinear up-sampling. This results in $\mathbf{f} \in \mathbb{R}^D$ where $D$ is the feature dimension and changes depending on the vision model $\phi$.

\noindent\textbf{Light-shadow, Size, and Texture-grad.} For these tasks, we calculate the difference between the masked average features of objects $A$ and $B$:
\begin{equation}
    \mathbf{f} = \frac{\sum_{h,w} M_A\odot \mathrm{up}(\phi(I))}{\sum_{h,w} M_A} - \frac{\sum_{h,w} M_B\odot \mathrm{up}(\phi(I))}{\sum_{h,w} M_B}. \label{eqn:feat_mask2}
\end{equation}
\noindent\textbf{Elevation and Perspective}. These tasks are not object/region specific, so we use the entire image feature, \ie $\mathbf{f}=\phi(I)$ $\in\mathbb{R}^{H_\mathbf{f}\times W_\mathbf{f}\times D}$, where $H_\mathbf{f}, W_\mathbf{f}$ are the height and width of the feature map.

\subsection{Probing Features}
\label{sec:protocol_probes}

For light-shadow, occlusion, size, and texture-grad, where the task-specific features $\mathbf{f}$ are $D$-dimensional vectors, we implement the probe $g_\theta$ as a multi-layer perception (MLP) with two layers and an intermediate GELU activation~\cite{hendrycks2016gaussian}. Though linear probing has been used commonly for probing large vision models~\cite{oquab2024dinov}, it is not obvious that the solutions to our tasks should be a linear function of the model features. Thus, we use non-linear probes, which empirically outperform linear probes (see \cref{supp:compare_probes}). For elevation and perspective, which require global information, we use an attentive probe~\cite{bardes2024vjepa,el2024scalable}. This probe aggregates global features through cross-attention and uses a two-layer MLP for final predictions. We observed via empirical results that the attentive probe achieves better performance than a standard MLP probe on these tasks.

\noindent\textbf{Training and Optimization.} We train the MLP probes for light-shadow, occlusion, size and texture-grad tasks using binary cross-entropy loss, as these tasks involve binary classification. For elevation and perspective, we use mean squared error loss to train the attentive probes, as the tasks are formulated as regression problems. %
Additional details on the probe optimization are provided in \cref{supp:implementation_probing}.

Following~\cite{oquab2024dinov, zhan2024physd}, we perform a hyper-parameter search on the layers of model $\phi$. For each task, we extract features from different layers of $\phi$ which is frozen, and train a probe on these features under the same setting. The best layer is then selected based on the validation performance, and the probing on this layer is repeated five times using different random seeds. 
We report the mean and standard deviation of the test performance.

\subsection{Evaluation Metrics}
For light-shadow, occlusion, size and texture-grad which are binary tasks, we report classification accuracy. For the vanishing point estimation task of the perspective cue, while Euclidean distance (between the predicted and ground-truth vanishing points) is a sensible metric, we instead report success rate to align this metric more closely with accuracy. 
Specifically, if the Euclidean distance of a prediction falls below a pre-defined threshold, it is interpreted as success. 
For elevation, we convert the horizon detection error~\cite{workman2016hlw} to an accuracy-based measure using a similar approach. %
More details are provided in \cref{supp:implementation_metrics}.

\section{Experiments}

\begin{figure*}
    \centering
    \includegraphics[width=.99\linewidth]{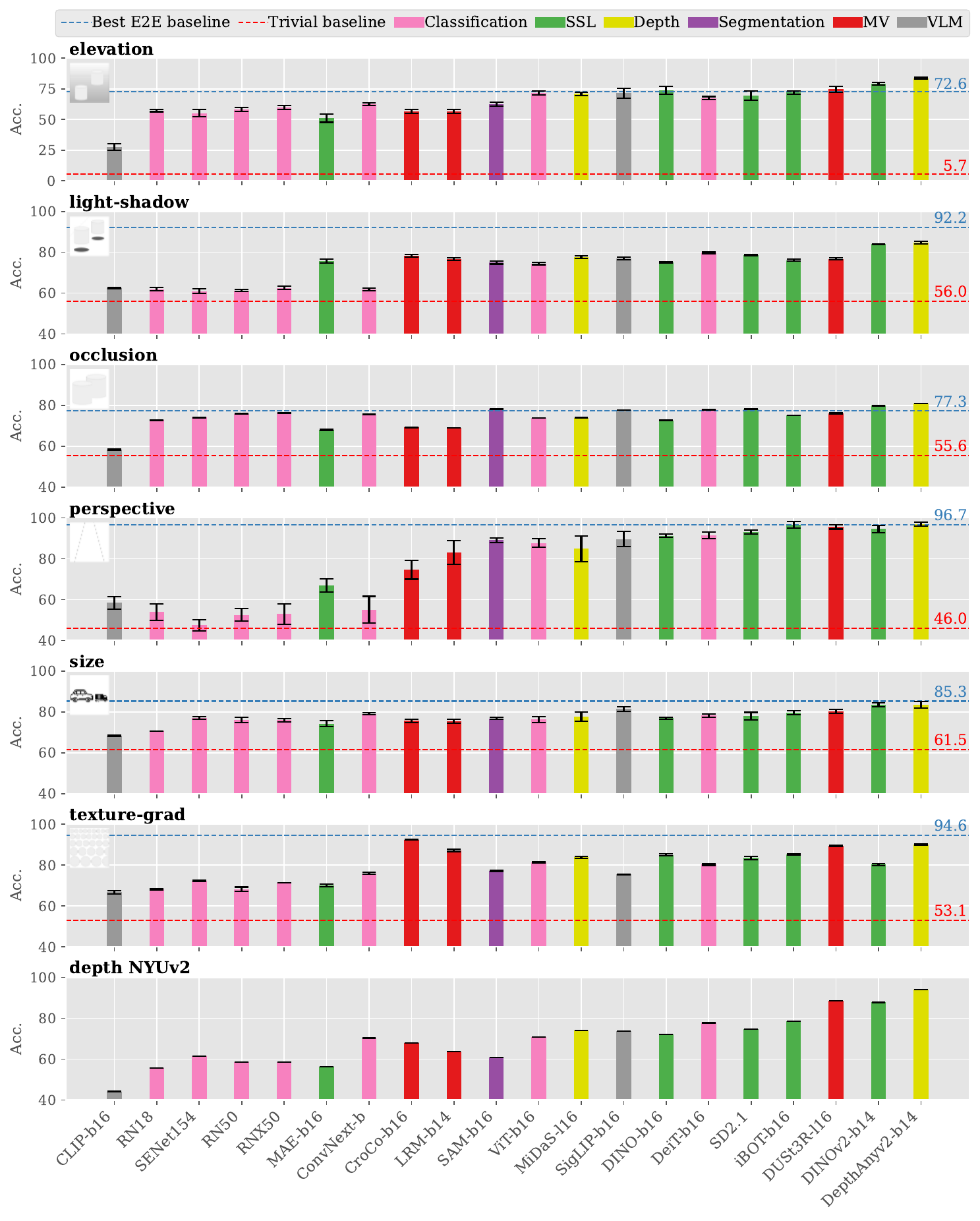}
    \vspace{-14pt}
    \caption{ 
        \textbf{\name Benchmark Results.} We evaluate 20 vision models with diverse pre-training settings (indicated by color) on the \name benchmark, which assesses six different monocular depth cues (each row) ubiquitous to humans. The models are ranked based on their average performance on the six cues. We include an end-to-end trained baseline (blue dotted line) as an oracle and a trivial baseline (red dotted line) to mark floor performance. Additionally, depth estimation linear probing results on NYUv2 are shown on the bottom row. 
    } 
    \label{fig:probing_results}
    \vspace{-12pt}
\end{figure*}

\subsection{Monocular Depth Cues in Vision Models}
\label{sec:exp_probing}

We evaluate 20 pre-trained large vision models~\cite{he2016deep,xie2017aggregated,hu2018squeeze,dosovitskiy2020vit,touvron2022deit,liu2022convnet,he2022masked,zhou2021ibot,caron2021emerging,oquab2024dinov,rombach2022high,ranftl2020towards,depth_anything_v2,hong2024lrm,weinzaepfel2022croco,wang2024dust3r,kirillov2023segment,radford2021learning,zhai2023sigmoid} spanning diverse pre-training objectives, modalities, and model architectures to assess their understanding of monocular depth perception cues in our \name benchmark.

\noindent\textbf{Implementation Details.} Where possible, we focus on model architectures equivalent to ViT-Base and denote the architecture size (and patch size for transformers) using postfixes, \eg DINOv2-b14. %
More details about the models and their architecture are provided in \cref{supp:selected_models}.
For consistency, we use only patch tokens for all ViT-based models~\cite{dosovitskiy2020vit}, as some do not include a class token.

\noindent{\bf Additional Baselines.} During the probing of certain cues, we explicitly use mask information to construct the input feature, $\mathbf{f}$. Although these masks are not directly used by the probes, they could theoretically be leveraged to solve tasks. To verify that any additional information from this step is insufficient to solve the task, we designed a \textit{trivial baseline} in which model features are replaced with simple 2D coordinate maps \ie $\phi(I)$ in \cref{eqn:feat_mask1,eqn:feat_mask2} becomes $C\in\mathbb{R}^{H\times W\times 2}$ where $C_{h,w} = [h,w]^T$. Moreover, for each cue, we include an \textit{end-to-end} baseline, where we train ResNet18~\cite{he2016deep} and ViT-B/16~\cite{dosovitskiy2020vit} models from scratch in an end-to-end fashion and select the best. We concatenate the masks with the input images channel-wise, allowing us to assess how much of each task can be solved without relying on sophisticated pre-trained features. Finally, to confirm that the tasks are readily solvable by humans, we evaluate our own performance on a random sample from the test sets, and obtained an accuracy of 95\%$\pm$1.48\% across tasks.

\noindent\textbf{Additional Tasks.} To extend our analysis, we also evaluate vision models for monocular depth estimation (on NYUv2~\cite{silberman2012indoor} and DIW~\cite{chen2016single}) and image classification (on ImageNet-1k~\cite{deng2009imagenet}) using a linear probe, a standard approach in previous work~\cite{oquab2024dinov, yue2024improving, aygun2024enhancing} for probing these datasets. Following the previous works~\cite{bhat2021adabins, chen2016single, el2024probing}, 
For NYUv2, we report accuracy as the percentage of pixels where the ratio between predicted and ground-truth depth is less than 1.25. 
We report top-1 accuracy for ImageNet. 
More details on these tasks are provided %
in \cref{supp:full_results},
and we show the linear probing results of these models on NYUv2 in \cref{fig:probing_results}.

\begin{figure}[t]
    \centering
    \includegraphics[width=\linewidth]{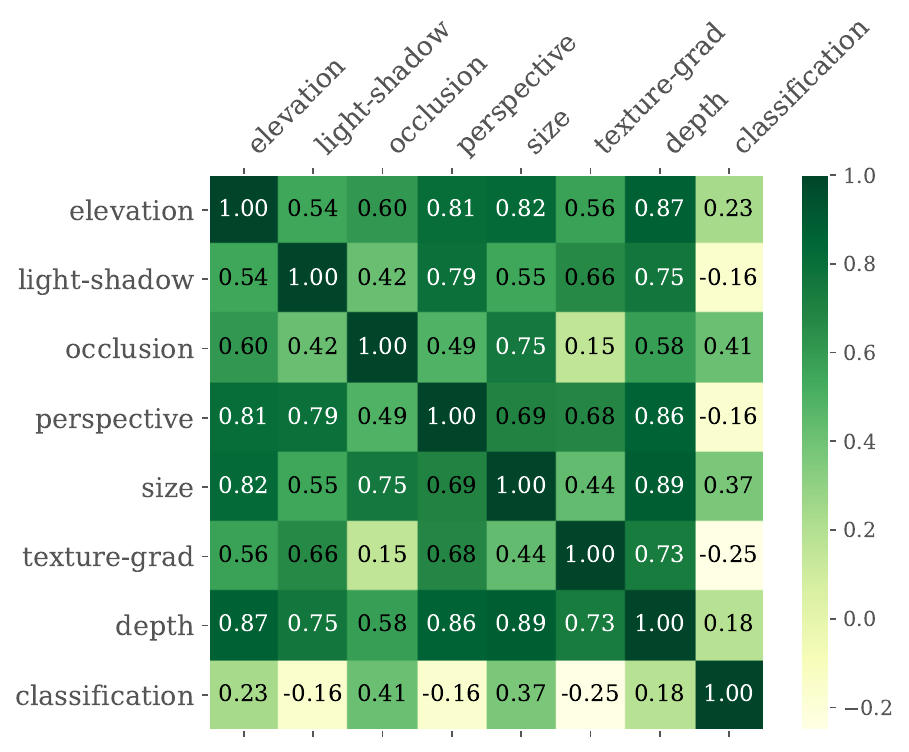}
    \vspace{-20pt}
    \caption{
        \textbf{Task correlation.} We measure the Spearman Ranked-order Correlation between each pair of tasks in \name benchmark, and how they correlate with depth estimation and image classification performance of models. A correlation score of one indicates the same ranking of models for two tasks.
    }
    \label{fig:task_correlation}
    \vspace{-14pt}
\end{figure}

\noindent\textbf{Results.} The probing results of 20 pre-trained vision models and the additional baselines on our \name benchmark are summarized in ~\cref{fig:probing_results}, where the models are ranked by average performance over all cues. Moreover, we measured Spearman Ranked-order Correlation between each pair of tasks in \name using the ranks of the models evaluated, and also how they correlate with the depth estimation and image classification performance of models, which is visualized in ~\cref{fig:task_correlation}. We discuss our observations below, and present additional analysis, such as how pre-training data size impacts performance, %
in \cref{supp:full_results}.

\noindent\textbf{(i) Depth cues emerge in more recent large vision models.} Firstly, the specialized depth estimation model, DepthAnythingv2, achieved the highest overall performance on \name, indicating the possibility that it leverages such cues to estimate depth.
We also observe the emergence of human-like monocular depth cues in recent large vision models, such as DINOv2 and Stable Diffusion. 
Notably, newer large vision models demonstrate a stronger grasp of these depth cues, as illustrated in \cref{fig:performance_vs_year}.
DINOv2 and Stable Diffusion have proven useful for developing monocular depth estimation models~\cite{depth_anything_v2,ke2024repurposing}. 
However, the reasons behind these models' effectiveness for depth inference remain unclear, given the lack of explicit depth supervision during pre-training.
Our results provide an insight on the development of depth perception in these models: they develop an awareness of human-like monocular depth cues, which might aid depth inference.

\noindent\textbf{(ii) Depth models do not always have the strongest knowledge of depth cues.} While DepthAnythingv2 achieves the best overall performance on \name, another specialized depth estimation model, MiDaS, is outperformed by several models pre-trained without depth supervision, \eg DeiT, SigLIP, and DINO. 
These models also exhibit competitive depth estimation performance (under linear probing) to MiDaS. This suggests that other pre-training recipes than dense depth supervision can effectively enhance depth perception, although they may be less data-efficient -- for instance, SigLIP is trained on around 10,000 times more data than MiDaS.

\noindent\textbf{(iii) Vision-language models exhibit varying levels of depth cue understanding.} CLIP obtains the lowest performance on \name, explaining its limited depth estimation ability as seen in \cref{fig:probing_results} and in recent work~\cite{el2024probing}. 
CLIP's limited understanding of the visual concepts tested in \name may present challenges for downstream tasks that require geometric understanding~\cite{zhu2024llava}. In contrast, SigLIP, also trained on matching images to captions but on significantly more data 
($\times9$ more data than CLIP), has a significantly better grasp of the depth cues. %

\noindent\textbf{(iv) Multi-view models are better at the texture-grad task than single-view models.} On all depth cues except texture-grad, recent models which are pre-trained with single-view images showed overall better performance than multi-view ones. However, on texture-grad, all three multi-view models (\ie CroCo, LRM, and DUSt3R) ranked within the top four, outperforming strong SSL competitors such as DINOv2. 
While the other five cues operate at the scene or object level, texture-grad uniquely requires a more localized, surface-level understanding. Multi-view pre-trained models are often more effective at capturing such details than single-view models because, during pre-training, they are exposed to variations in the same surface across different viewpoints. For instance, CroCo, which achieves the best performance on the texture-grad task, is trained to reconstruct parts of a scene from different viewpoints, requiring attention to finer local details. Moreover, LRM and DUSt3R are pre-trained on 3D object and scene reconstruction tasks, which require understanding correspondences across views by leveraging local surface details.

\noindent\textbf{(v) Performance on \name correlates with depth estimation performance.}
As shown in \cref{fig:task_correlation}, all \name tasks exhibit moderate to high levels of correlation with depth estimation, supporting the importance of depth cues for this task. However, all tasks have a low correlation with ImageNet-1k classification. Interestingly, a relatively higher alignment with classification is observed for occlusion and size, which are also highly correlated with each other.
This is reasonable as understanding occlusion and size requires object-level semantics (\eg `What does the recognized object look like when not occluded?' and `Is category A larger than category B?'). 
Apart from texture-grad, each cue has a high correlation ($>0.75$) with at least one other cue. As discussed above, the texture-grad cue requires understanding surface details and provides localized depth information; perception of this cue is found to be weaker in single-view image models. We note that texture-grad is the only cue that uses synthetic data, which might also contribute to its lower correlation with other cues.

\subsection{Learning Depth Cues}
\label{sec:method_finetuning}

\begin{table}[t]
\centering
\resizebox{\linewidth}{!}{
\begin{tabular}{l cc}
\toprule
Model                                          & NYUv2 Acc. (\%) $\uparrow$     & DIW WHDR (\%) $\downarrow$  \\ \midrule
DINOv2                                         & 87.78                          & 11.99 \\
DINOv2+\textit{DC}                             & 87.06                          & 11.95 \\
\texttt{concat}(DINOv2, noise)                 & 87.56                          & 12.20 \\
\texttt{concat}(DINOv2, DINOv2+\textit{DC} )   & \textbf{88.46}                 & \textbf{11.72} \\ 
\midrule
CLIP                                           & 43.78                          & 35.25 \\
CLIP+\textit{DC}                               & 43.59                          & 35.45 \\
\texttt{concat}(CLIP, noise)                   & 43.38                          & 35.39 \\
\texttt{concat}(CLIP, CLIP+\textit{DC} )       & \textbf{44.32}                 & \textbf{33.53} \\
\bottomrule
\end{tabular}
}
\vspace{-5pt}
\caption{
    \textbf{Linear probing on downstream depth estimation.} For each model, we probe the patch tokens from the final layer. Models with `+\textit{DC}' are fine-tuned using supervision on our \name benchmark.
    The best results in each subset are \textbf{bolded}.
}
\vspace{-10pt}
\label{tab:downstream_depth}
\end{table}

We observed that the depth estimation performance of a model highly correlates with its awareness of human-like monocular depth cues. Here we ask if enhanced depth cue awareness improves depth estimation? To investigate this, we fine-tune single-image pre-trained vision models on the training set of \name in a multi-task manner, then evaluate the fine-tuned model on downstream depth estimation. Here we focus on two models widely adopted for various downstream applications: DINOv2 and CLIP.

We use low-rank adaptation (LoRA)~\cite{hu2022lora} for efficient fine-tuning of pre-trained models. Further details of multi-task learning and LoRA implementation are provided %
in \cref{supp:implementation_finetune}.
We evaluate the pre-trained and fine-tuned models on NYUv2~\cite{silberman2012indoor} and DIW~\cite{chen2016single} depth datasets using linear probes. 
We experiment with two different variations: first, we evaluate the features from fine-tuned models (denoted as +\textit{DC} in \cref{tab:downstream_depth}); second, we concatenate the features from fine-tuned models with the features from the original pre-trained models. The latter approach was also adopted in~\cite{yue2024improving, mariotti2024improving} to preserve the generalization ability of the pre-trained models.
As before, we report accuracy for NYUv2, and use Weighted Human Disagreement Rate (WHDR) to measure performance on DIW. %
It is noted the dataset for the size cue in \name contains a subset of NYUv2 images (due to the use of SUNRGBD), so evaluation on DIW offers a fairer comparison. %

We observe in \cref{tab:downstream_depth} that for both models, fine-tuning on \name resulted in improved depth perception. Notably, the fine-tuned features often need to be concatenated with the original features to bring improvements, a similar observation also made in previous works~\cite{yue2024improving, mariotti2024improving}. A possible explanation for the decreased generalization in the models is due to the limited size and diversity of the fine-tuning data in \name.
We also investigate whether the improvements in the concatenated variations come from the increased parameters in the probe (due to higher dimensional input), by training linear probes on a combination of original model features and random noise (denoted as \texttt{concat}($\cdot$, noise)), which did not result in improvements.

\subsection{Discussion}
The results in \cref{sec:exp_probing} indicate that recent large vision models understand several human-like monocular depth cues, with performance on these cues correlating with depth estimation performance. This suggests that models may use such cues to infer depth, even without explicit depth supervision. For example, the popular self-supervised model, DINOv2, has developed features that help reason about these cues (except texture-grad), which can support depth estimation.
Interestingly, the results in \cref{sec:method_finetuning} show that \name enables the learning of depth cues, which benefit downstream depth performance in a linear probing setting, even though \name has a moderate size (fewer than 35k training images, much smaller than the large datasets used to pre-train the base models), is not designed specifically for end-to-end fine-tuning, and only provides image-level sparse annotations (as opposed to dense depth maps). This suggests promising directions for enhancing depth estimation from sparse annotations.

\noindent\textbf{Limitations.} 
Our analysis focuses only on widely used pre-trained model weights but not directly on the impact of various pre-training objectives and their training datasets on the emergence of depth cues. While depth from ego-motion and scene motion are also important cues for humans, our focus is primarily on assessing monocular cues learned by large vision models trained on static image collections and not image sequences (except CroCo, LRM, and DUSt3R).

\section{Conclusions}
We introduced \name, a benchmark suite for evaluating human-like monocular depth cues in large vision models. Testing 20 diverse models, we found that depth perception cues, similar to those used by humans, emerge across various pre-training objectives, with more recent and larger models demonstrating stronger alignment with human depth perception. Furthermore, our fine-tuning experiments on \name indicate that a model’s depth perception can be enhanced by explicitly incorporating human-like depth priors, suggesting promising directions for improving depth performance in future models. We hope \name will serve as a valuable resource for advancing research in depth-aware vision systems, particularly for applications requiring robust 3D spatial understanding from 2D inputs. 

\noindent\textbf{Acknowledgments.} Funding was provided by ELIAI (the Edinburgh Laboratory for Integrated Artificial Intelligence) - EPSRC (EP/W002876/1). HB was supported by the EPSRC Visual AI grant EP/T028572/1, and CJ was supported by a gift from Adobe.

\clearpage
{
    \small
    \bibliographystyle{ieeenat_fullname}
    \bibliography{main}
}

\clearpage
\appendix 

\noindent{\LARGE \bf Appendix}

\setcounter{table}{0}
\renewcommand{\thetable}{A\arabic{table}}
\setcounter{figure}{0}
\renewcommand{\thefigure}{A\arabic{figure}}

\input{supp_content}

\end{document}

%% file: supp_content.tex
\section{Additional Results on \textit{DepthCues}}
\label{supp:additional_results}

\subsection{Full Benchmark Results} \label{supp:full_results}
In the main paper, we reported the performance of 20 pre-trained vision models on \name and NYUv2~\cite{silberman2012indoor} using bar charts, and focused on salient observations. Here we provide detailed numerical results on these datasets and additionally the linear probing results on ImageNet-1k~\cite{deng2009imagenet}. 
Moreover, we also provide an analysis of the impact of the pre-training dataset, objective, and architectural choices of the vision models on their performance. Our analysis is based solely on publicly available models; therefore, as outlined in the limitations section of the main paper, we are unable to control all variables in these analyses.

\begin{figure}[t]
    \centering
    \includegraphics[width=\linewidth]{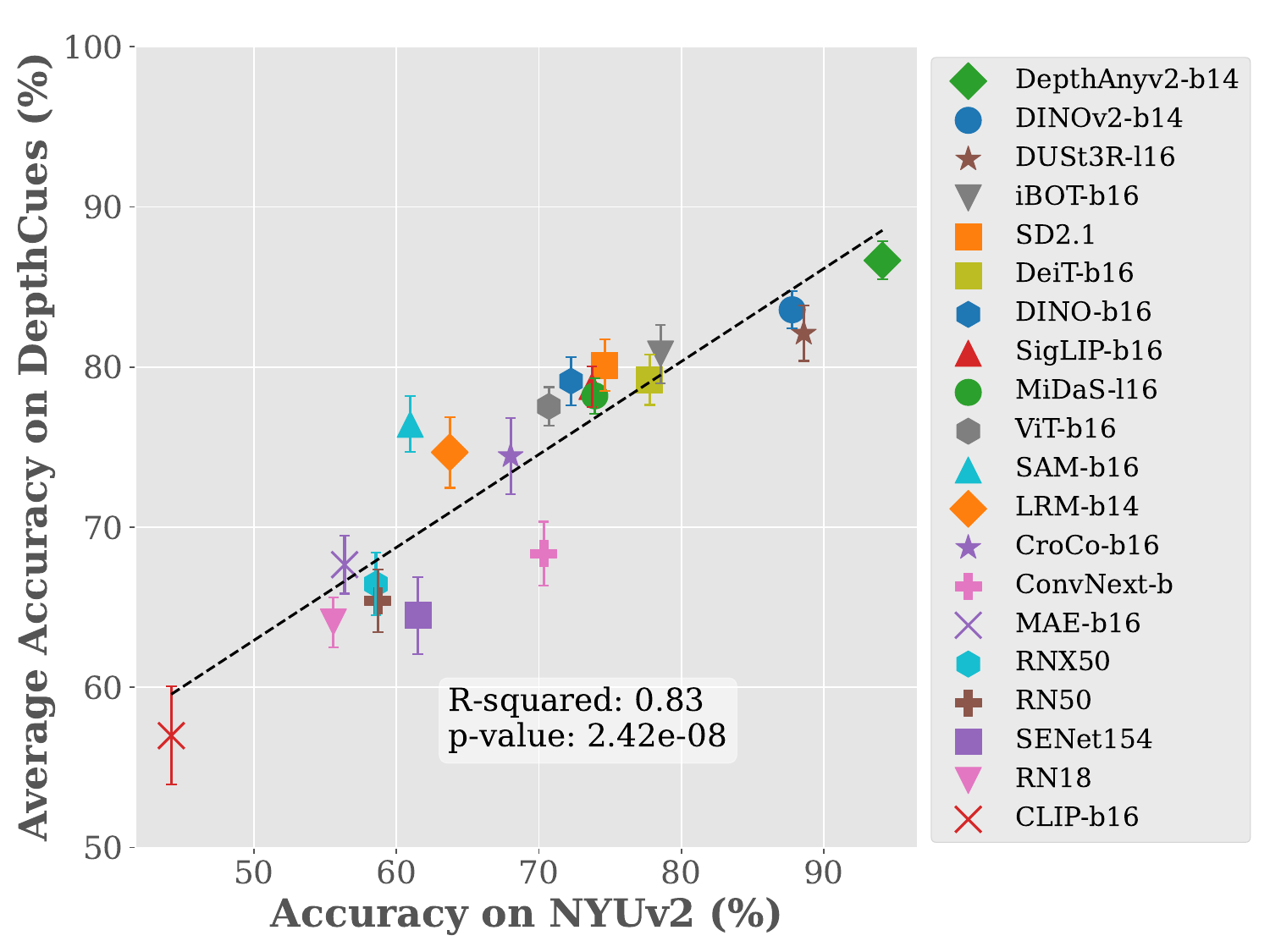}
    \vspace{-18pt}
    \caption{ 
        \textbf{Performance of vision models on \name vs. NYUv2 depth estimation.} A strong correlation is observed between depth cue understanding and depth estimation.
    }
    \label{fig:performance_vs_nyuv2}
\end{figure}

\begin{figure}[t]
    \centering
    \includegraphics[width=\linewidth]{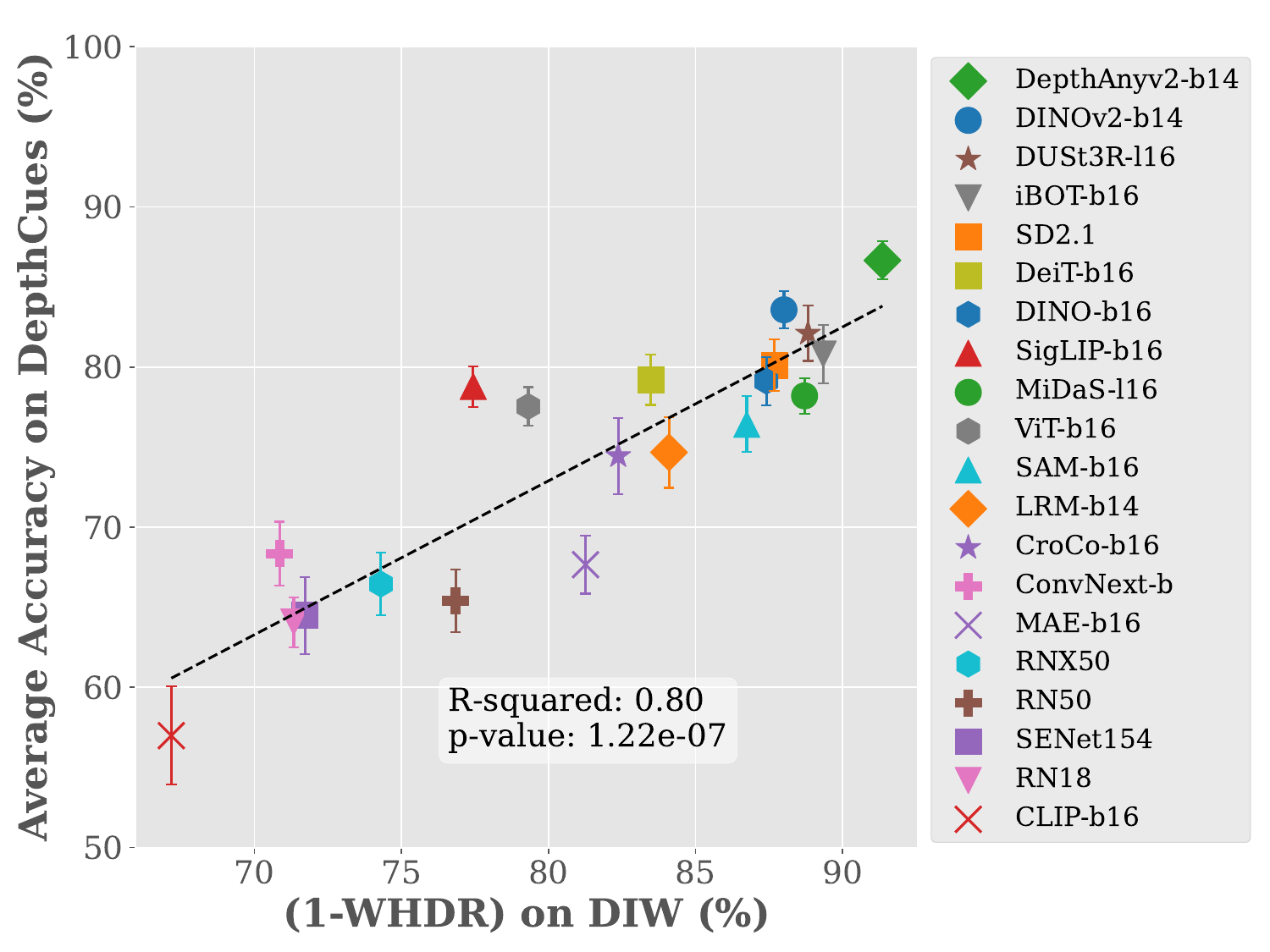}
    \vspace{-18pt}
    \caption{ 
        \textbf{Performance of vision models on \name vs. DIW depth ordering.} A strong correlation is observed between depth cue understanding and depth estimation.
    }
    \label{fig:performance_vs_diw}
\end{figure}

The probing results of all models are summarized in \cref{tab:fullresults}, and the average performance of models on \name is plotted against their NYUv2 depth estimation (\cref{fig:performance_vs_nyuv2}) as well as DIW depth ordering (\cref{fig:performance_vs_diw}) results. It is noted that there are slight differences between the linear probing results on NYUv2 and ImageNet-1k compared to previous works~\cite{oquab2024dinov, el2024probing}. These may stem from differences in the implementation details including the features probed, learning rates, and training iterations, \etc. For example, the probing results on NYUv2 depth estimation in \cite{el2024probing} are generally higher than our results. Possible reasons for this include the following factors. 
Firstly, while we train a simple linear layer to probe the patch tokens from the last layer of the vision models, a more complex non-linear probe similar to the DPT decoder~\cite{ranftl2021vision} is used to probe features from multiple layers of the models in~\cite{el2024probing}. Secondly, the class tokens are utilized for several models (\eg DINO, DINOv2) in \cite{el2024probing}, while we only use patch tokens for consistent evaluation across all models as some of the models do not include a class token. Thirdly, because of the difference in the probes, their optimization hyper-parameters can differ. Similar reasons can be applied to the differences in our ImageNet-1k probing results from previous works. In particular, for CLIP, the class token seems to be crucial for its ImageNet classification ability. We made these design choices because our primary focus is to evaluate the awareness of human-like monocular depth cues in vision models in a fair and comparable manner, rather than optimizing their probing performance on external benchmarks. 

\begin{figure}[t]
    \centering
    \includegraphics[width=\linewidth]{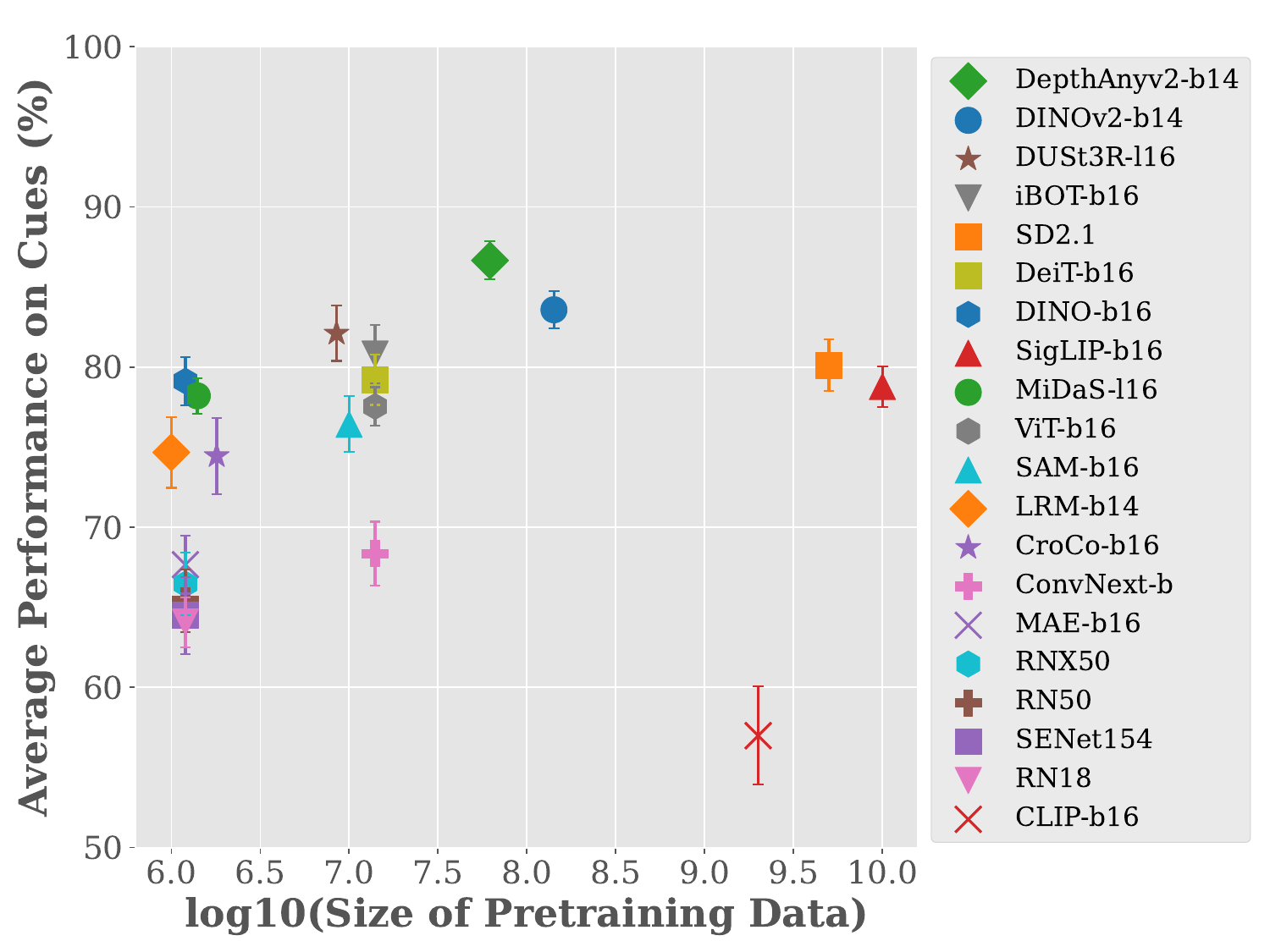}
    \vspace{-18pt}
    \caption{ 
        \textbf{Impact of pre-training dataset size on the performance of vision models on \name.} Omitting the three models that involve language supervision (CLIP, SigLIP, and SD), we observe a moderate positive correlation between depth cue understanding and pre-training data size.
    }
    \label{fig:performance_vs_data}
\end{figure}

\begin{table*}[t]
\centering
\resizebox{\linewidth}{!}{
\begin{tabular}{rcccccccc}
\toprule
\multicolumn{1}{l}{} & \multicolumn{6}{c}{\textit{DepthCues}}                                                                       & NYUv2      & ImageNet-1k         \\
\cmidrule(l{5pt}r{5pt}){2-7}\cmidrule(l{5pt}r{5pt}){8-8}\cmidrule(l{5pt}r{5pt}){9-9}
               & elevation (\%) & light-shadow (\%) & occlusion (\%) & perspective (\%) & size (\%)    & texture-grad (\%) & depth (\%) & classification (\%) \\
\midrule
CLIP-b16       & 27.56 (2.76)   & 62.49 (0.19)      & 58.41 (0.20)   & 58.40 (3.06)     & 68.37 (0.21) & 66.68 (0.77)      & 44.20      & 0.59                \\
RN18           & 57.17 (0.83)   & 61.86 (0.84)      & 72.79 (0.20)   & 53.87 (3.97)     & 70.57 (0.11) & 68.22 (0.35)      & 55.57      & 68.67               \\
SENet154       & 55.19 (2.97)   & 60.95 (1.04)      & 74.07 (0.09)   & 47.47 (2.71)     & 76.94 (0.63) & 72.30 (0.32)      & 61.51      & 81.58               \\
RN50           & 58.38 (1.61)   & 61.26 (0.48)      & 75.96 (0.18)   & 52.53 (3.05)     & 76.01 (1.35) & 68.26 (1.00)      & 58.70      & 73.62               \\
RNX50          & 59.72 (1.73)   & 62.49 (0.82)      & 76.36 (0.20)   & 52.93 (5.07)     & 75.93 (0.77) & 71.28 (0.07)      & 58.56      & 76.40               \\
MAE-b16        & 51.03 (3.33)   & 75.72 (0.99)      & 68.05 (0.22)   & 66.93 (3.23)     & 74.27 (1.40) & 69.96 (0.66)      & 56.36      & 25.99               \\
ConvNext-b     & 62.52 (1.21)   & 61.86 (0.64)      & 75.60 (0.22)   & 55.07 (6.52)     & 79.04 (0.47) & 75.98 (0.50)      & 70.34      & 80.37               \\
CroCo-b16      & 56.64 (1.57)   & 78.28 (0.66)      & 69.21 (0.08)   & 74.53 (4.57)     & 75.64 (0.72) & 92.40 (0.14)      & 68.02      & 14.52               \\
LRM-b14        & 56.58 (1.75)   & 76.56 (0.68)      & 69.14 (0.02)   & 83.07 (5.79)     & 75.51 (0.97) & 87.18 (0.51)      & 63.75      & 17.68               \\
SAM-b16        & 62.33 (1.56)   & 74.93 (0.70)      & 78.29 (0.10)   & 89.07 (1.16)     & 76.74 (0.49) & 77.18 (0.43)      & 60.97      & 19.58               \\
ViT-b16        & 71.68 (1.57)   & 74.42 (0.50)      & 73.94 (0.06)   & 87.60 (2.09)     & 76.29 (1.42) & 81.28 (0.23)      & 70.70      & 68.67               \\
MiDaS-l16      & 71.04 (1.38)   & 77.70 (0.66)      & 74.02 (0.16)   & 84.93 (6.29)     & 77.70 (2.27) & 83.80 (0.50)      & 73.93      & 42.79               \\
SigLIP-b16     & 71.44 (4.12)   & 76.98 (0.53)      & 77.76 (0.08)   & 89.60 (3.64)     & 81.40 (1.10) & 75.52 (0.13)      & 73.73      & 37.96               \\
DINO-b16       & 73.80 (3.15)   & 75.19 (0.29)      & 72.63 (0.18)   & 91.20 (0.78)     & 76.88 (0.40) & 85.00 (0.54)      & 72.26      & 41.29               \\
DeiT-b16       & 67.44 (1.26)   & 79.81 (0.41)      & 78.00 (0.12)   & 91.47 (1.66)     & 78.17 (0.92) & 80.28 (0.37)      & 77.77      & 84.53               \\
SD2.1          & 69.45 (3.56)   & 78.53 (0.26)      & 78.20 (0.14)   & 93.07 (1.00)     & 77.92 (1.85) & 83.50 (0.74)      & 74.63      & 32.67               \\
iBOT-b16       & 72.08 (1.32)   & 76.05 (0.50)      & 75.19 (0.09)   & 96.67 (1.58)     & 79.61 (0.89) & 85.24 (0.34)      & 78.54      & 38.92               \\
DUSt3R-l16     & 74.65 (2.44)   & 76.75 (0.45)      & 76.02 (0.30)   & 95.47 (0.98)     & 80.31 (0.92) & 89.42 (0.35)      & 88.59      & 25.94               \\
DINOv2-b14     & 79.13 (1.11)   & 83.95 (0.25)      & 79.93 (0.17)   & 94.53 (1.86)     & 83.57 (0.93) & 80.32 (0.49)      & 87.78      & 77.95               \\
DepthAnyv2-b14 & 83.74 (0.57)   & 84.74 (0.68)      & 81.01 (0.08)   & 96.93 (0.90)     & 83.51 (1.59) & 89.98 (0.34)      & 94.12      & 68.67 \\
\bottomrule
\end{tabular}
}
\vspace{-5pt}
\caption{
    \textbf{Detailed evaluation results on \name, depth estimation (NYUv2), and image classification (ImageNet-1k).} The mean and standard deviation of the accuracy of each vision model on the six tasks in our benchmark are summarized. The last two columns show the accuracy of these models on NYUv2 depth estimation and ImageNet-1k classification.
}
\label{tab:fullresults}
\end{table*}

\begin{figure}[t]
    \centering
    \includegraphics[width=\linewidth]{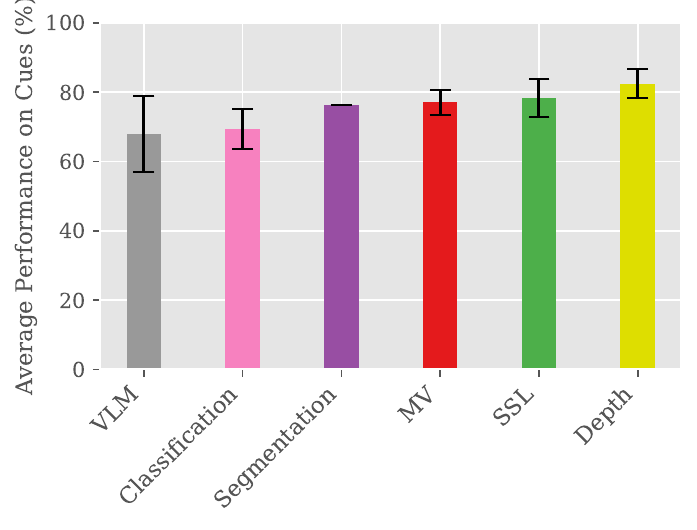}
    \vspace{-18pt}
    \caption{ 
        \textbf{Average performance of vision models on \name by pre-training objective.} The error bar denotes the standard deviation over models. 
    }
    \label{fig:performance_vs_objective}
\end{figure}

\begin{figure}[t]
    \centering
    \subfloat[]{\includegraphics[width=0.34675\linewidth]{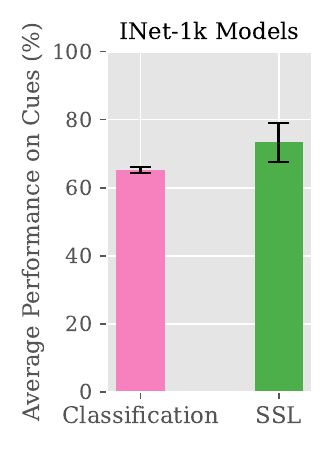}}\;\!\!
    \subfloat[]{\includegraphics[width=0.3016\linewidth]{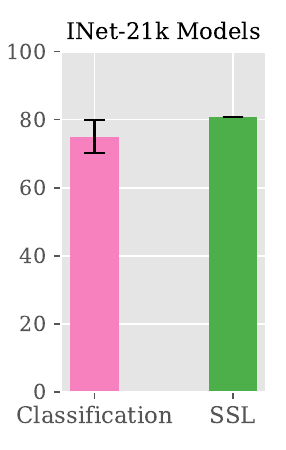}}\;\!\!
    \subfloat[]{\includegraphics[width=0.30875\linewidth]{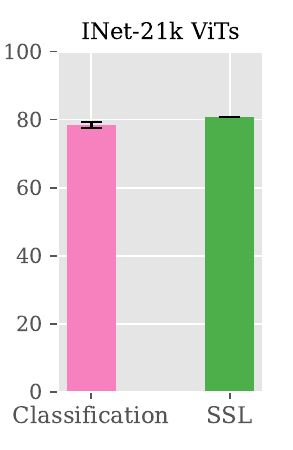}}
    \vspace{-8pt}
    \caption{ 
        \textbf{Average performance of vision models on \name by pre-training objective, with fixed pre-training data.} We show results for ImageNet-1k (a) and ImageNet-21k models (b), and ImageNet-21k ViT-based models (c). The error bar denotes the standard deviation over models. 
    }
    \label{fig:performance_vs_objective_imagenet}
\end{figure}

\begin{figure}[t]
    \centering
    \subfloat[]{\includegraphics[width=0.5035\linewidth]{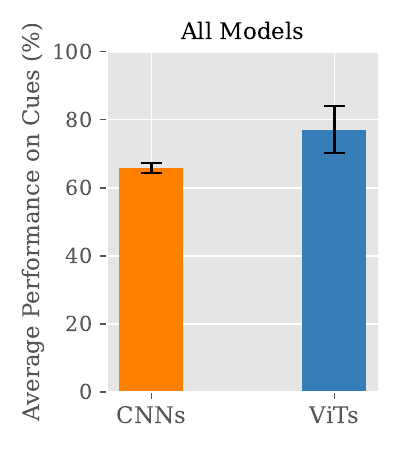}}\;\!\!
    \subfloat[]{\includegraphics[width=0.4464\linewidth]{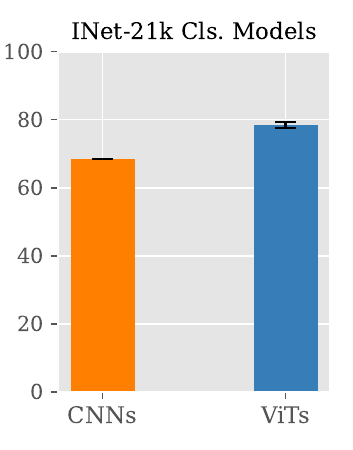}}\;\!\!
    \vspace{-8pt}
    \caption{ 
        \textbf{Average performance of vision models on \name by model architecture.} We show results for all models (a) and ImageNet-21k classification models (b). The error bar denotes the standard deviation over models. 
    }
    \label{fig:performance_vs_arch}
\end{figure}

\begin{figure}[t]
    \centering
    \includegraphics[width=0.78\linewidth]{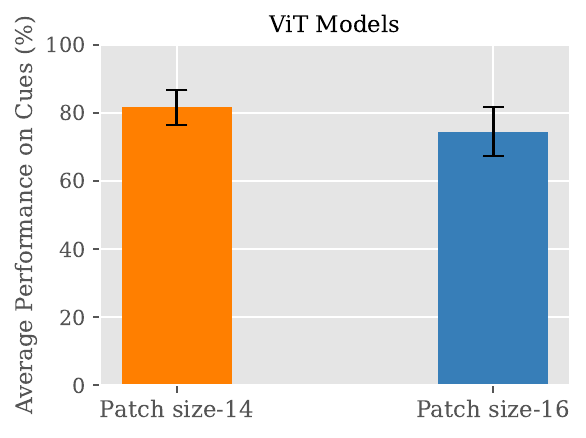}
    \vspace{-8pt}
    \caption{ 
        \textbf{Average performance of ViT-based models on \name by patch size.} The error bar denotes the standard deviation over models. 
    }
    \vspace{-8pt}
    \label{fig:performance_vs_patchsize}
\end{figure}

\noindent\textbf{Impact of Pre-Training Dataset Size.} \cref{fig:performance_vs_data} shows the average performance of models on \name against the size of their pre-training datasets. Excluding two vision-language models (CLIP and SigLIP) and the generative model (StableDiffusion, which also trained for text-to-image generation using language supervision), there is a moderate positive correlation between these two variables (Pearson's $r=0.69, p=0.002$). Apart from the overall trend, we observe that certain models demonstrate greater data efficiency in developing depth cue awareness. For instance, among models pre-trained on approximately $10^6$ data samples, DINO achieves the best performance. For those pre-trained on around $10^7$ samples, DUSt3R emerges as the best. Finally, when comparing models pre-trained on roughly one order of magnitude more data, DepthAnythingv2 outperforms DINOv2.
However, there are other factors such as the pre-training objective, model architecture, and the dataset distribution which can confound these analyses. It is desirable to perform future studies on the pre-training dataset size where the aforementioned factors are better controlled.

\noindent\textbf{Impact of Pre-Training Objective.} We report the average performance achieved on \name by models pre-trained with different types of supervision in \cref{fig:performance_vs_objective}. On average, models pre-trained for depth estimation obtain the highest performance, followed by (single-view) self-supervised (SSL) and multi-view models. Considering the differences in the pre-training data of these groups, we also collate results for different groups when the pre-training dataset is the same. The only models that support such analysis are either classification or SSL-based (more details in \cref{supp:selected_models}), and their results are shown in \cref{fig:performance_vs_objective_imagenet}. We can see from \cref{fig:performance_vs_objective_imagenet}~(a) and (b) that SSL methods have developed a better understanding of studied depth cues on average. However, in these two plots, the architecture of classification models is mixed (convolutional and transformer-based models), whereas SSL models only contain ViT backbones. Therefore, we further reduce the effect of architecture and show a comparison for ViT-only models (`Base' models; patch size 16) in \cref{fig:performance_vs_objective_imagenet}~(c), where we still see an advantage from the SSL model (iBOT) over the classification ones (ViT and DeiT).

\begin{figure}[t]
    \centering
    \includegraphics[width=\linewidth]{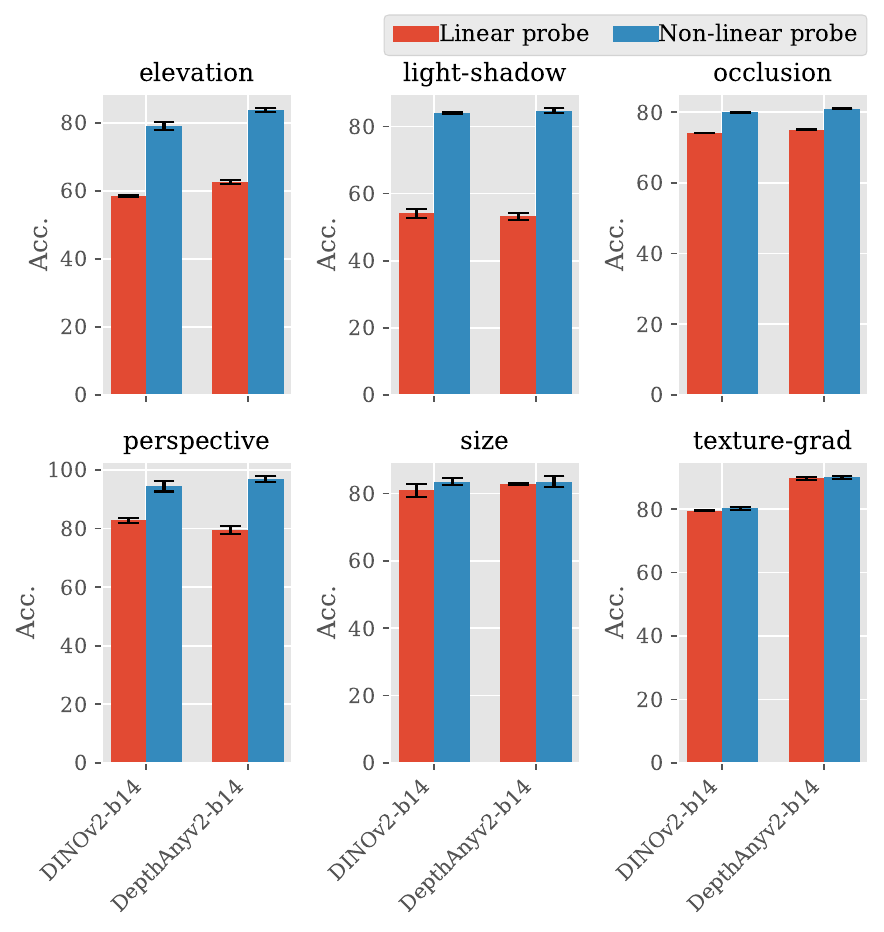}
    \vspace{-18pt}
    \caption{ 
        \textbf{Linear vs. non-linear probes.} We see consistently better performance from non-linear probes across the tasks in \name.
    }
    \label{fig:probes_linear_vs_mlp}
\end{figure}

\begin{figure}[t]
    \centering
    \includegraphics[width=\linewidth]{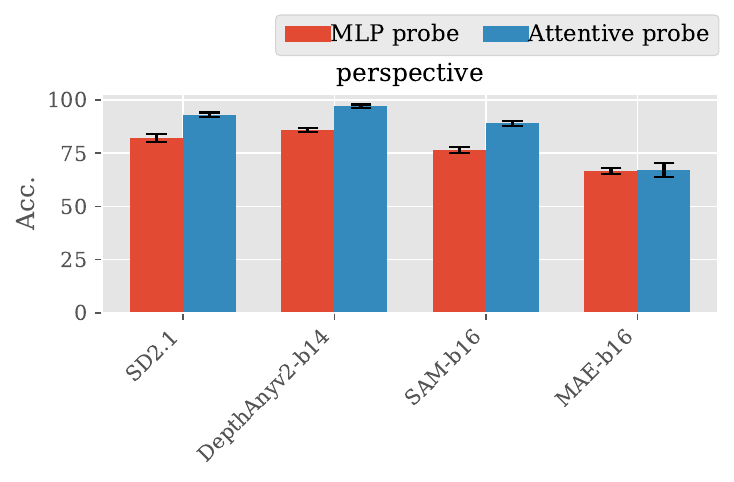}
    \vspace{-18pt}
    \caption{ 
        \textbf{MLP vs. Attentive probe on the perspective task.} We observe that for a task that requires global information, the attentive probe yields better results than an MLP. 
    }
    \vspace{-8pt}
    \label{fig:probes_mlp_vs_attn}
\end{figure}

\noindent\textbf{Impact of Model Architecture.} Here we compare the average depth cue awareness between vision transformer (ViT)-based models and models with convolutional architectures (CNNs). \cref{fig:performance_vs_arch}~(a) shows the results for all evaluated models, regardless of pre-training objective and dataset, where we see a clear advantage of the ViT models. To reduce confounding factors, in \cref{fig:performance_vs_arch}~(b) we also compare these architectures when the pre-training setting is fixed to be classification on ImageNet-21k. It is observed that under such a setting, the two ViT models (ViT and DeiT) demonstrate enhanced knowledge of depth cues compared to ConvNext.

\noindent\textbf{Impact of Patch Size on ViT Models.} We also perform a comparison between ViT models regarding their patch size. As shown in \cref{fig:performance_vs_patchsize}, models with a smaller patch size (14 \vs 16) achieved better average performance on \name. However, it is noted that the ``patch size-14'' models only include DINOv2, DepthAnythingv2, and LRM, and there may be other factors that led to their superior performance.

\begin{table*}[t]
\centering
\resizebox{\linewidth}{!}{
\begin{tabular}{rrcccccccccccc}
\toprule
\multirow{2}{*}{Model} & \multirow{2}{*}{Layers Probed} & \multicolumn{4}{c}{Elevation (\%)}                                     & \multicolumn{4}{c}{Light-shadow (\%)}                                      & \multicolumn{4}{c}{Occlusion (\%)}                                         \\
\cmidrule(l{5pt}r{5pt}){3-6}\cmidrule(l{5pt}r{5pt}){7-10}\cmidrule(l{5pt}r{5pt}){11-14}
                       &                                & Block 1        & Block 2        & Block 3        & Block 4        & Block 1         & Block 2         & Block 3         & Block 4         & Block 1         & Block 2         & Block 3         & Block 4         \\
\midrule
CLIP-b16               & \{3, 6, 9, 12\}                & \textbf{0.189} & 0.243          & 0.372          & 0.355          & \textbf{66.357} & 64.965          & 65.893          & 65.429          & \textbf{58.283} & 57.531          & 55.912          & 56.206          \\
RN18                   & \{1, 2, 3, 4\}                 & 0.145          & 0.127          & 0.122          & \textbf{0.105} & 64.965          & 63.805          & \textbf{65.197} & 65.029          & 58.360          & 67.302          & \textbf{73.175} & 66.744          \\
SENet154               & \{1, 2, 3, 4\}                 & 0.120          & 0.108          & \textbf{0.105} & 0.112          & 65.893          & 66.357          & 66.473          & \textbf{67.749} & 60.987          & \textbf{74.066} & 73.532          & 68.681          \\
RN50                   & \{1, 2, 3, 4\}                 & 0.138          & 0.118          & \textbf{0.096} & 0.109          & 65.197          & 65.429          & 65.313          & \textbf{69.026} & 59.391          & 65.737          & \textbf{75.701} & 70.735          \\
RNX50                  & \{1, 2, 3, 4\}                 & 0.136          & 0.113          & \textbf{0.097} & 0.113          & 65.777          & 65.545          & 66.357          & \textbf{68.329} & 59.329          & 69.913          & \textbf{76.554} & 67.480          \\
MAE-b16                & \{3, 6, 9, 12\}                & 0.151          & 0.130          & \textbf{0.116} & 0.122          & 81.323          & 81.787          & \textbf{82.251} & 80.510          & 61.623          & 63.381          & \textbf{67.550} & 66.744          \\
ConvNext-b             & \{1, 2, 3, 4\}                 & 0.129          & 0.121          & 0.108          & \textbf{0.096} & 65.197          & 66.125          & \textbf{69.722} & 67.169          & 59.623          & 65.838          & \textbf{75.136} & 71.347          \\
CroCo-b16              & \{3, 6, 9, 12\}                & 0.125          & 0.111          & \textbf{0.107} & 0.108          & 83.411          & \textbf{83.991} & 83.759          & 83.411          & 62.870          & 65.946          & \textbf{68.712} & 68.271          \\
LRM-b14                & \{3, 6, 9, 12\}                & 0.135          & 0.108          & \textbf{0.108} & 0.114          & 68.677          & 81.903          & \textbf{84.919} & 83.411          & 62.142          & 67.651          & \textbf{69.650} & 67.256          \\
SAM-b16                & \{3, 6, 9, 12\}                & 0.133          & 0.111          & 0.085          & \textbf{0.084} & 80.046          & \textbf{80.858} & 80.046          & 80.278          & 63.513          & 72.571          & 78.080          & \textbf{78.700} \\
ViT-b16                & \{3, 6, 9, 12\}                & 0.101          & \textbf{0.081} & 0.087          & 0.098          & 79.466          & \textbf{81.671} & 76.798          & 70.186          & 64.637          & 72.393          & \textbf{73.291} & 67.883          \\
MiDaS-l16              & \{6, 12, 18, 24\}              & 0.108          & \textbf{0.078} & 0.091          & 0.083          & \textbf{83.759} & 80.974          & 79.814          & 78.886          & 69.882          & \textbf{73.524} & 73.408          & 73.222          \\
SigLIP-b16             & \{3, 6, 9, 12\}                & 0.090          & 0.072          & \textbf{0.072} & 0.083          & 80.858          & \textbf{86.543} & 81.903          & 71.694          & 68.852          & \textbf{76.809} & 74.508          & 70.277          \\
DINO-b16               & \{3, 6, 9, 12\}                & 0.099          & 0.073          & \textbf{0.073} & 0.074          & 81.206          & 82.483          & \textbf{83.179} & 81.903          & 65.791          & 72.641          & 73.470          & \textbf{73.617} \\
DeiT-b16               & \{3, 6, 9, 12\}                & 0.091          & \textbf{0.084} & 0.086          & 0.118          & 83.527          & \textbf{87.007} & 73.434          & 66.357          & 68.681          & \textbf{77.158} & 73.679          & 67.488          \\
SD2.1                  & \{1, 2, 3, 4\}                 & 0.100          & \textbf{0.094} & 0.096          & 0.139          & 81.090          & \textbf{88.399} & 79.698          & 66.125          & 69.371          & \textbf{77.096} & 72.129          & 59.468          \\
iBOT-b16               & \{3, 6, 9, 12\}                & 0.114          & 0.093          & 0.081          & \textbf{0.078} & 78.190          & 82.947          & 83.527          & \textbf{83.527} & 62.994          & 72.772          & \textbf{75.027} & 73.570          \\
DUSt3R-l16             & \{18, 24, 33, 36\}             & \textbf{0.066} & 0.076          & 0.083          & 0.093          & 83.759          & \textbf{86.427} & 85.383          & 82.251          & 73.764          & 69.975          & 74.330          & \textbf{75.833} \\
DINOv2-b14             & \{3, 6, 9, 12\}                & 0.117          & 0.073          & 0.068          & \textbf{0.065} & 67.053          & 79.234          & \textbf{89.211} & 87.935          & 63.412          & 76.399          & \textbf{79.544} & 75.205          \\
DepthAnyv2-b14         & \{3, 6, 9, 12\}                & 0.114          & 0.070          & 0.064          & \textbf{0.058} & 66.473          & 79.350          & \textbf{90.603} & 90.023          & 63.436          & 77.111          & \textbf{80.962} & 77.057          \\
\bottomrule
\end{tabular}
}
\resizebox{\linewidth}{!}{
\begin{tabular}{rrcccccccccccc}
\toprule
\multirow{2}{*}{Model} & \multirow{2}{*}{Layers Probed} & \multicolumn{4}{c}{Perspective (\%)}                                     & \multicolumn{4}{c}{Size (\%)}                                      & \multicolumn{4}{c}{Texture-grad (\%)}                                         \\
\cmidrule(l{5pt}r{5pt}){3-6}\cmidrule(l{5pt}r{5pt}){7-10}\cmidrule(l{5pt}r{5pt}){11-14}
                       &                                & Block 1        & Block 2        & Block 3        & Block 4        & Block 1         & Block 2         & Block 3         & Block 4         & Block 1         & Block 2         & Block 3         & Block 4         \\
\midrule
CLIP-b16               & \{3, 6, 9, 12\}                & 0.259          & \textbf{0.230} & 0.253          & 0.264          & \textbf{73.636} & 73.635          & 72.727          & 70.909          & \textbf{67.000} & 64.900          & 63.600          & 61.200          \\
RN18                   & \{1, 2, 3, 4\}                 & 0.290          & 0.265          & \textbf{0.235} & 0.259          & 68.485          & \textbf{78.788} & 77.273          & 77.273          & \textbf{67.400} & 65.300          & 64.500          & 63.700          \\
SENet154               & \{1, 2, 3, 4\}                 & 0.270          & \textbf{0.250} & 0.262          & 0.294          & 74.242          & 77.879          & \textbf{83.333} & 81.515          & 69.200          & \textbf{74.800} & 66.300          & 64.400          \\
RN50                   & \{1, 2, 3, 4\}                 & 0.296          & 0.290          & \textbf{0.230} & 0.264          & 73.030          & 75.455          & \textbf{78.788} & 78.485          & 68.500          & 69.000          & \textbf{70.200} & 67.300          \\
RNX50                  & \{1, 2, 3, 4\}                 & 0.294          & 0.270          & \textbf{0.240} & 0.277          & 70.606          & 78.182          & \textbf{80.606} & 76.061          & 68.900          & \textbf{72.600} & 69.000          & 65.200          \\
MAE-b16                & \{3, 6, 9, 12\}                & 0.264          & 0.240          & 0.218          & \textbf{0.217} & 76.364          & \textbf{79.697} & 79.696          & 78.485          & 66.300          & 69.500          & 68.900          & \textbf{69.800} \\
ConvNext-b             & \{1, 2, 3, 4\}                 & 0.291          & 0.296          & \textbf{0.244} & 0.248          & 71.818          & 74.242          & 80.303          & \textbf{81.818} & 70.800          & \textbf{73.600} & 61.600          & 63.600          \\
CroCo-b16              & \{3, 6, 9, 12\}                & 0.234          & 0.177          & \textbf{0.148} & 0.153          & 76.364          & 77.273          & \textbf{79.091} & 78.788          & 74.900          & 89.800          & 91.600          & \textbf{92.700} \\
LRM-b14                & \{3, 6, 9, 12\}                & 0.253          & 0.200          & 0.146          & \textbf{0.137} & 75.152          & 78.485          & \textbf{80.000} & 79.697          & 66.900          & 72.900          & 81.000          & \textbf{87.800} \\
SAM-b16                & \{3, 6, 9, 12\}                & 0.224          & 0.178          & \textbf{0.116} & 0.145          & 76.667          & 80.909          & 80.606          & \textbf{80.909} & 76.800          & \textbf{78.200} & 74.600          & 71.800          \\
ViT-b16                & \{3, 6, 9, 12\}                & 0.149          & 0.109          & \textbf{0.108} & 0.213          & 77.879          & 79.394          & \textbf{80.000} & 78.788          & 75.100          & \textbf{82.600} & 76.000          & 70.100          \\
MiDaS-l16              & \{6, 12, 18, 24\}              & 0.227          & \textbf{0.135} & 0.137          & 0.158          & 76.970          & \textbf{78.788} & 77.879          & 78.485          & 84.100          & \textbf{85.200} & 84.000          & 83.300          \\
SigLIP-b16             & \{3, 6, 9, 12\}                & 0.181          & \textbf{0.111} & 0.124          & 0.161          & 79.091          & 81.515          & 79.697          & \textbf{82.121} & 76.700          & \textbf{78.300} & 69.700          & 61.700          \\
DINO-b16               & \{3, 6, 9, 12\}                & 0.152          & \textbf{0.078} & 0.092          & 0.115          & 77.576          & \textbf{80.303} & 77.576          & 79.091          & 80.000          & 84.700          & \textbf{84.900} & 84.000          \\
DeiT-b16               & \{3, 6, 9, 12\}                & 0.179          & \textbf{0.107} & 0.194          & 0.251          & 77.576          & \textbf{80.909} & 80.303          & 80.606          & 79.900          & \textbf{81.100} & 72.100          & 62.500          \\
SD2.1                  & \{1, 2, 3, 4\}                 & \textbf{0.102} & 0.110          & 0.105          & 0.297          & 79.091          & \textbf{81.212} & 79.091          & 73.636          & 81.800          & \textbf{83.200} & 79.600          & 70.300          \\
iBOT-b16               & \{3, 6, 9, 12\}                & 0.182          & 0.086          & \textbf{0.069} & 0.078          & 79.091          & 80.909          & 79.697          & \textbf{81.818} & 80.400          & 83.600          & \textbf{85.100} & 84.200          \\
DUSt3R-l16             & \{18, 24, 33, 36\}             & \textbf{0.076} & 0.097          & 0.098          & 0.106          & 83.636          & \textbf{85.455} & 82.121          & 77.576          & 87.500          & 87.600          & 91.700          & \textbf{91.900} \\
DINOv2-b14             & \{3, 6, 9, 12\}                & 0.218          & 0.142          & 0.089          & \textbf{0.088} & 79.091          & 79.697          & 82.727          & \textbf{85.455} & 77.200          & \textbf{82.500} & 80.800          & 80.600          \\
DepthAnyv2-b14         & \{3, 6, 9, 12\}                & 0.189          & 0.091          & \textbf{0.085} & 0.092          & 77.576          & 83.030          & 82.424          & \textbf{86.667} & 76.700          & 83.000          & 86.100          & \textbf{90.000}          \\
\bottomrule
\end{tabular}
}
\caption{
    \textbf{Layer search for all models on all tasks in \name.} We report the validation performance of different layers, and \textbf{bold} the best score for each model. Note that ``Block $i$'' corresponds to the $i$th layer indicated in the second column. Horizon detection error and Euclidean distance are used to assess validation performance for elevation and perspective respectively, while accuracy is used for the other four tasks.
}
\label{tab:hp}
\end{table*}

\subsection{Comparison Between Probes}\label{supp:compare_probes}
As discussed in Sec.~4.2 in the main paper, we adopt non-linear probes to evaluate models on \name. Specifically, an MLP probe is used for light-shadow, occlusion, size, and texture-grad, and an attentive probe~\cite{bardes2024vjepa,el2024scalable} is used for elevation and perspective. 
Our motivation for using these instead of linear probes is that it is not clear that the solutions to our tasks should be a linear function of the model features. In addition, non-linear probes have been adopted in previous work on probing vision models for depth estimation~\cite{el2024probing,ge2024geobench}.
To justify our choice empirically, we compare our non-linear probes and the linear probe. Following the same protocol described in the main paper, we obtained additional linear probing results of two models (DINOv2 and DepthAnythingv2) on all six depth cue tasks. The results are summarized in \cref{fig:probes_linear_vs_mlp}, showing that our non-linear probes consistently outperform linear probes, based on the average of five runs, although the gap is small for size and texture-grad. Moreover, for the light-shadow task, the performance of the linear probe is close to random, indicating that a linear classifier is not sufficient to solve the problem using these models' features.

For elevation and perspective, our choice of the attentive probe, instead of the MLP probe used for the other four tasks, is motivated by the extra step in the former for aggregating global information, which we consider important for these two tasks. This is also supported by our results in \cref{fig:probes_mlp_vs_attn}, where it can be seen that the attentive probe results in significant performance gains for most models on the perspective task.

\begin{figure*}[t]
    \includeinkscape[width=1.0\textwidth]{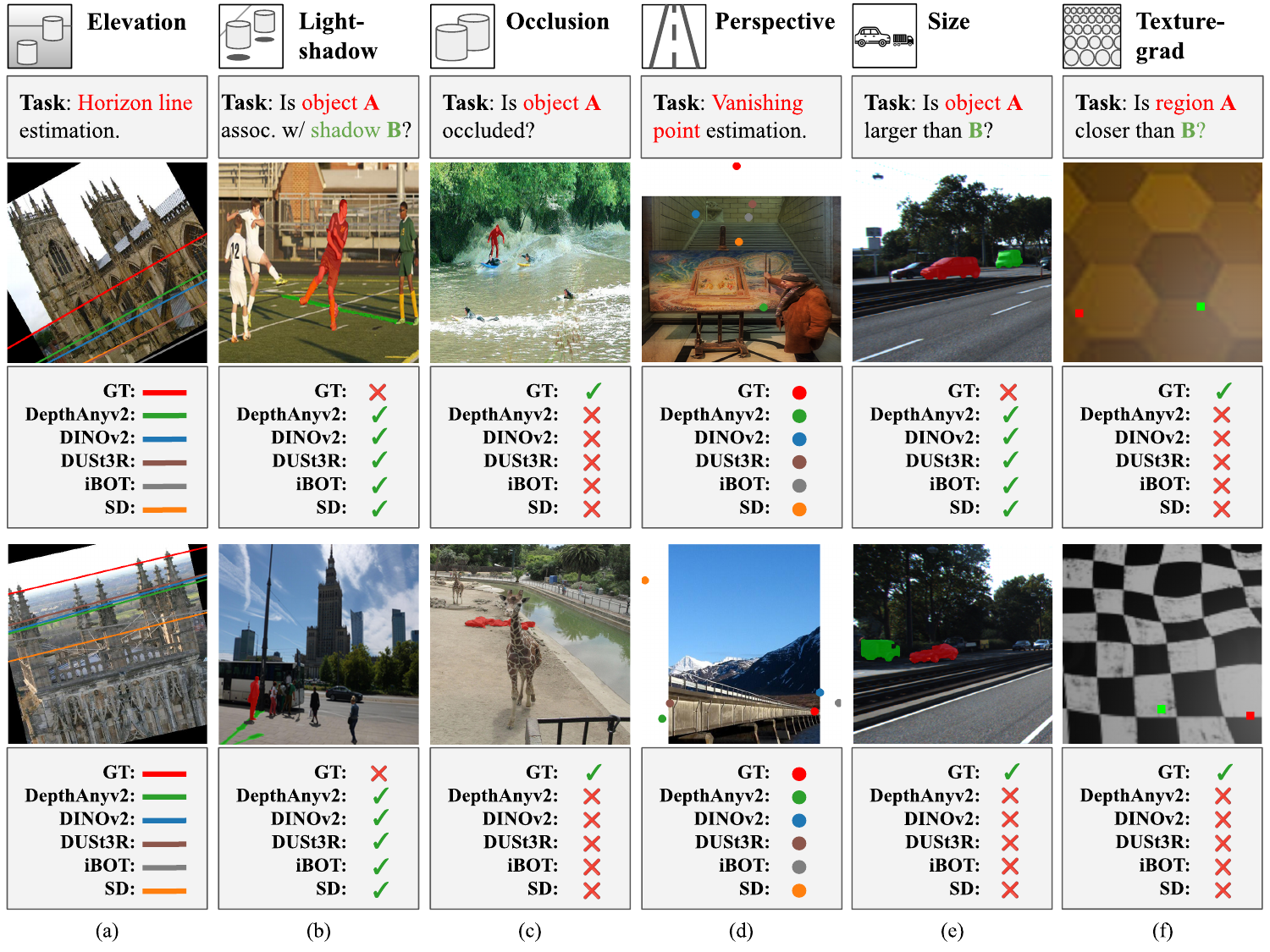}
    \vspace{-17pt}
    \caption{
        \textbf{Failure cases of the top-five vision models on \name.} Each column shows two examples for a depth cue.
    }
    \label{fig:failure_cases}
\end{figure*}

\subsection{Hyper-Parameter Search Results}
It has been shown in previous works~\cite{zhan2024physd, ma2024imagenetd} that different layers of pre-trained vision models have varied performance when probed for different tasks. Therefore, for all the 20 vision models, we perform a hyper-parameter search on their layers. %
We restrict the search to four layers for each model, which are selected by equally dividing the networks into four blocks (similar to \cite{el2024probing}), where applicable. We train the probes on features from each block, and evaluate their validation performance. The layer with the best validation result is then selected for subsequent analysis. The layer search results for all models are summarized in \cref{tab:hp}. Consistent with previous findings, we observe that different layers of a model exhibit varying strengths. For instance, in DINOv2, the 9th layer achieved the best performance on light-shadow and occlusion, the 12th layer provides superior features for elevation, perspective, and size, while texture-grad is handled best by the 6th layer.

\begin{figure*}[t]
    \centering
    \includegraphics[width=\linewidth]{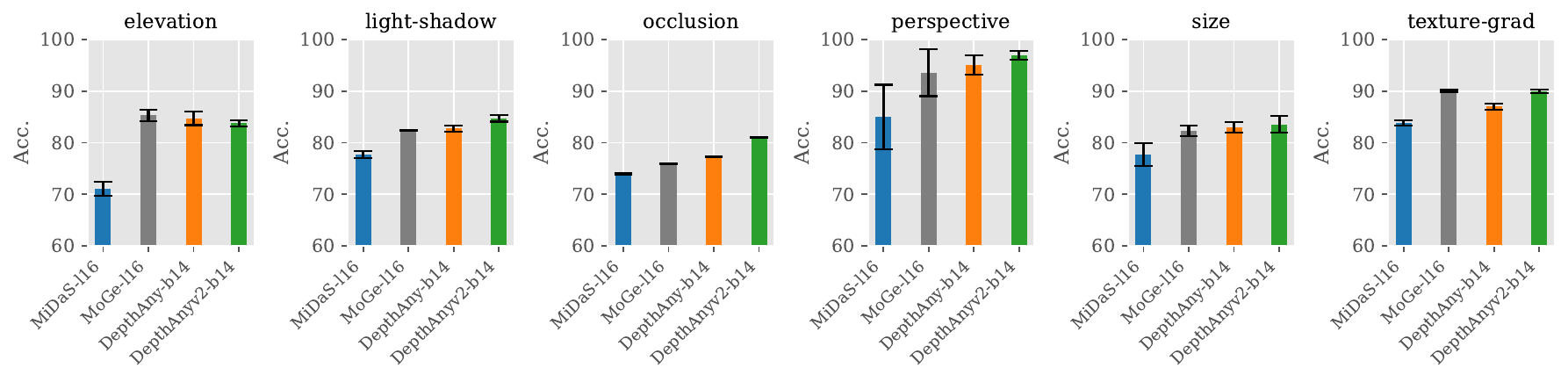}
    \vspace{-20pt}
    \caption{ 
        \textbf{Performance of monocular depth estimation models on \name.} Here we evaluate recent depth estimation models, and observe that these models show competitive performance, indicating their strong understanding of the monocular depth cues.
    }
    \label{fig:bar_depth_only}
\end{figure*}

\begin{figure}
    \centering
    \includegraphics[width=\linewidth]{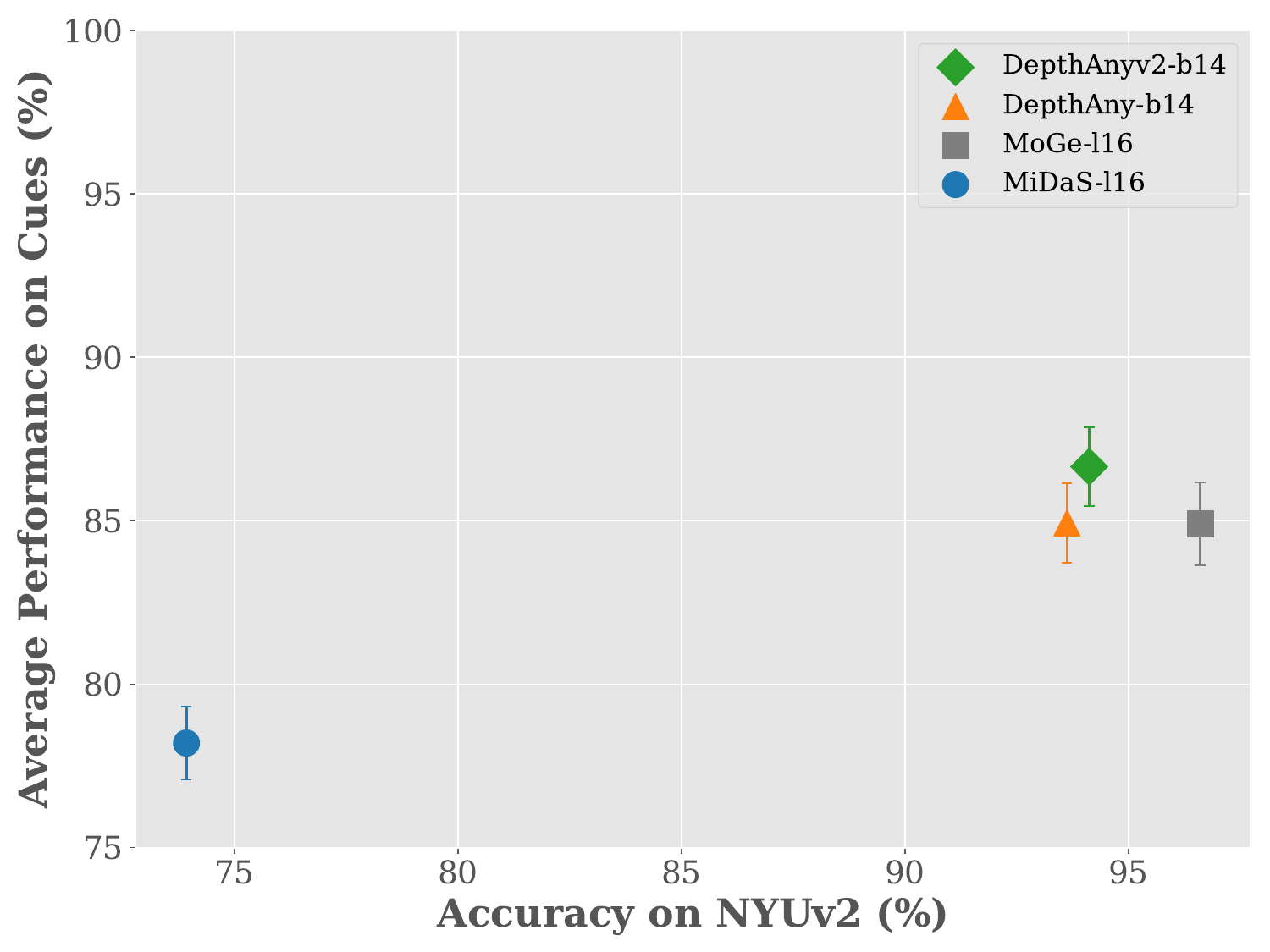}
    \vspace{-18pt}
    \caption{ 
        \textbf{Performance on monocular depth estimation models \name vs. NYUv2 depth estimation}.
    }
    \label{fig:performance_vs_nyuv2_depth_only}
\end{figure}

\subsection{Example Model Failure Cases}
We examine where vision models consistently fail. Focusing on the top-five models, namely DepthAnythingv2, DINOv2, DUSt3R, iBOT, and StableDiffusion, we identify test instances where they all fail to predict the correct label over five runs, and visualize these cases in \cref{fig:failure_cases}.

\noindent\textbf{Elevation.} In the first (top row) instance of \cref{fig:failure_cases}~(a), the horizon line is largely occluded by the architectures, making it a challenging case. In the second case (bottom row), although the horizon line is less occluded by objects, its visibility is still low due to the fog, which can be a possible explanation for the failure of the models.

\noindent\textbf{Light-shadow.} In both of the examples in \cref{fig:failure_cases}~(b), the target objects (overlaid with red masks) and query shadows (overlaid with green masks) overlap with each other, possibly affecting the models' predictions. This is especially the case for the second image, where the true shadow of the target object also overlaps with the false query shadow.

\noindent\textbf{Occlusion.} The first example (top row) of \cref{fig:failure_cases}~(c) is challenging because the person highlighted with a red mask undergoes only very minor occlusion, limited to his feet. In the second example, the stones are treated as one object and labeled occluded in the source data~\cite{zhu2017semantic}, which can be confused with their individual boundaries or occlusions.

\noindent\textbf{Perspective.} Both of the examples in \cref{fig:failure_cases}~(d) contain distractors that weaken the notion of the dominant vanishing point. In the first example (top), while the ground-truth label indicates the converging point of the two sides of the staircase, some models are influenced by the parallel lines formed by the bricks on the walls. In the second example (bottom), some models might be affected by the intersection line of the mountain and sky.

\noindent\textbf{Size.} In the first example of \cref{fig:failure_cases}~(e), all models incorrectly predicted that the car (highlighted in red) has a larger 3D size than the van (highlighted in green). In the second example, while the trailer (green) has a smaller size than the car, all models predicted the opposite case. These indicate limitations in the models' understanding of the size cue.

\noindent\textbf{Texture-grad.} All models failed to predict the correct depth order in both cases in \cref{fig:failure_cases}~(f). In these challenging images, the texture gradient cue is relatively weak, and the locations of the two regions are not drastically different, which can make these cases challenging.

\subsection{Learning Depth Cues}

\begin{table}[t]
\centering
\resizebox{\linewidth}{!}{
\begin{tabular}{lllllll}
\toprule
\multicolumn{1}{l}{} & \multicolumn{6}{c}{\textit{DepthCues} (\%)}                                                                             \\
\cmidrule(l{5pt}r{5pt}){2-7}
               & elevation  & light-shadow  & occlusion  & perspective  & size   & texture-grad \\
\midrule
DINOv2             & 77.46 & 83.26 & 75.35 & 96.00 & 83.57 & 78.90 \\
DINOv2+\textit{DC} & 79.84 (+2.38) & 82.67 (-0.59) & 76.11(+0.76) & 94.00 (-2.00) & 86.10 (+2.53) & 88.30 (+9.40) \\
\bottomrule
\end{tabular}
}
\vspace{-5pt}
\caption{
    \textbf{Probing results of the original and fine-tuned DINOv2 on \name.} Here `+\textit{DC}' indicates the fine-tuned model. We observe an overall increase in performance, especially on texture-grad.
}
\label{tab:probe_finetuned}
\end{table}

\begin{table}[t]
\centering
\resizebox{\linewidth}{!}{
\begin{tabular}{l cc}
\toprule
Model                                          & NYUv2 Acc. (\%) $\uparrow$     & DIW WHDR (\%) $\downarrow$  \\ \midrule
DINOv2                                         & 87.81 (0.09)                   & 11.95 (0.05) \\
\texttt{concat}(DINOv2, DINOv2+\textit{DC} )   & 88.43 (0.09)                   & 11.66 (0.11) \\ 
\midrule
CLIP                                           & 43.82 (0.04)                   & 35.33 (0.09) \\
\texttt{concat}(CLIP, CLIP+\textit{DC} )       & 44.40 (0.18)                   & 32.69 (0.98) \\
\bottomrule
\end{tabular}
}
\vspace{-5pt}
\caption{
    \textbf{Linear probing on downstream depth estimation with different random seeds.} Fine-tuning on \name benchmark is also repeated.
}
\vspace{-7pt}
\label{tab:downstream_depth_repeat}
\end{table}

In Sec.~5.2 of the main paper, we presented linear probing results on NYUv2 and DIW, showing that fine-tuning on \name improves models' performance on depth estimation. Here we further investigate whether the fine-tuned model has improved understanding of the depth cues by comparing the probing results of the fine-tuned  and pre-trained DINOv2 on \name. Note we focus on probing the last layer since LoRA was only applied on that layer during fine-tuning. The results from \cref{tab:probe_finetuned} shows that the fine-tuned version outperforms the original DINOv2 on four cues, with a slight drop in performance on light-shadow and perspective. Notably, we see a 9.4\% increase in accuracy on texture-grad, on which the original DINOv2 shows relatively weaker understanding (see Fig.~3 in the main paper). In addition, we repeated the fine-tuning (on \name) and evaluation of DINOv2 and CLIP on NYUv2 and DIW with three different random seeds, and show the results in \cref{tab:downstream_depth_repeat}, where significant improvements are observed.

\subsection{Monocular Depth Estimation Models}
In addition to the 20 vision models evaluated in the main paper, here we present additional results for some recent monocular depth estimation models: DepthAnything~\cite{depth_anything_v1} and MoGe~\cite{wang2024moge}. The average performance of these models on \name and depth estimation is shown in \cref{fig:bar_depth_only} and \cref{fig:performance_vs_nyuv2_depth_only}, along with the two depth models already included in our study. In general, we observe a similar trend to our previous findings, that the models' depth estimation performance highly correlates with their performance on \name. 

\noindent\textbf{Relative vs. Metric Depth Models.} The DepthAnythingv2 model checkpoint we evaluated so far was pre-trained for relative depth estimation. Here we additionally evaluate the two publicly available metric versions of DepthAnythingv2 on \name, which are fine-tuned from the relative version on indoor and outdoor datasets respectively. It is observed from \cref{tab:rebuttal_metric} that the metric versions achieved lower performance than the relative one. As metric models are fine-tuned from the relative one on smaller datasets, the difference in performance could also be impacted by the training data. We leave further investigation to future work.

\begin{table}[t]
\centering
\resizebox{\linewidth}{!}{
\begin{tabular}{lllllll}
\toprule
DAv2 ver.      & Ele. & Lit-shd. & Occ. & Prsp. & Size  & Txt.-grd. \\
\midrule
relative       & 83.73     & 84.65        & 81.16     & 95.33       & 82.16 & 89.90        \\
abs. (outdoor) & 74.96     & 82.91        & 78.60     & 92.00       & 83.85 & 88.70        \\
abs. (indoor)  & 79.96     & 82.44        & 79.67     & 94.67       & 83.43 & 89.90       \\
\bottomrule
\end{tabular}
}
\vspace{-5pt}
\caption{
    \textbf{Probing results of relative \vs metric versions of DepthAnyv2 on \name}.
}
\vspace{-5pt}
\label{tab:rebuttal_metric}
\end{table}

\begin{figure*}[ht]
    \includeinkscape[width=1.0\textwidth]{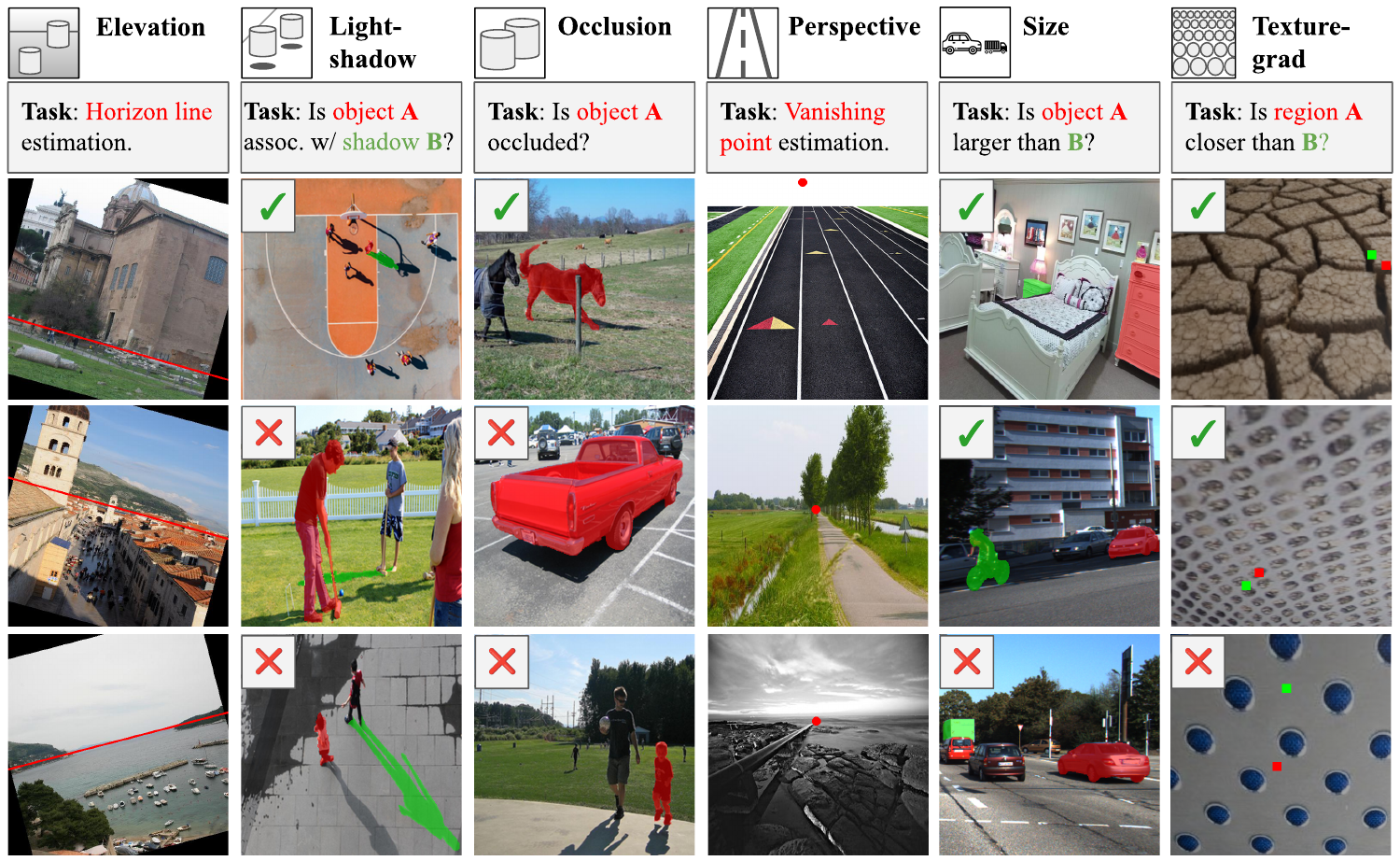}
    \vspace{-17pt}
    \caption{
        \textbf{Examples data instances from \name.} The horizon lines for elevation and the vanishing points for perspective are indicated by red lines and dots respectively. For the other four cues where we define binary classification tasks, the labels are indicated by the ticks and crosses.
    }
    \label{fig:more_examples}
    \vspace{-8pt}
\end{figure*}

\section{Dataset Construction Details}
\label{supp:dataset}

Here we provide additional details on the construction of the \name benchmark and show other examples from the datasets.

\noindent\textbf{Instance Selection for Size.} To create an instance for the size dataset, we sample two objects from an image, and obtain the label by comparing the volumes of their 3D bounding boxes (provided by the source datasets, KITTI~\cite{geiger2012we} and SUN-RGBD~\cite{song2015sun}). It is mentioned in the main paper that a threshold is applied to filter out cases where the difference in the sizes of the two objects is very small. This is motivated by our observation that the 3D bounding boxes from the source datasets can contain minor errors. Therefore, to reduce mislabeling in the size dataset, we apply a minimum threshold of 2.5 $m^3$ for the size difference between two objects in an image from KITTI, and a threshold of 0.4 $m^3$ for SUN-RGBD. We found empirically that these thresholds provided a good balance between label accuracy and dataset size. 

\noindent\textbf{Generating Masks with SAM.} Four of the tasks in \name, namely light-shadow, occlusion, size, and texture-grad, require object masks for the probing evaluation. The object masks for light-shadow and occlusion are directly obtained from their source datasets~\cite{Wang_2020_CVPR,zhu2017semantic}, and the masks for texture-grad are manually defined during dataset synthesis (see Sec.~3.6 in the main paper). However, the object masks for the size task are not available in the source datasets which are originally designed for 3D object detection. Therefore, we made use of an off-the-shelf segmentation model, SAM~\cite{kirillov2023segment}, to create these masks. Specifically, for each object, we obtain its 2D bounding box from the source dataset, and discard the object if its 2D bounding box has a height/width less than eight pixels, to filter out potentially incorrect annotations. Next, we predict the mask for an object by feeding its 2D bounding box and the image to SAM, and only keep the part of the mask that falls into the 2D bounding box. Finally, to make sure the desired object is segmented, we check whether the mask takes up a too small portion of the bounding box ($<20\%$), and discard the object if that is the case. These filtering steps are applied to both objects in a candidate image, and the candidate is included in the dataset only if both object masks pass these checks (and satisfy the size difference threshold specified above).

\noindent\textbf{More Examples from the Datasets.} Additional examples from the \name datasets are shown in \cref{fig:more_examples}.

\section{Additional Implementation Details}
\label{supp:implementation}

\begin{figure}[t]
    \includeinkscape[width=1.0\linewidth]{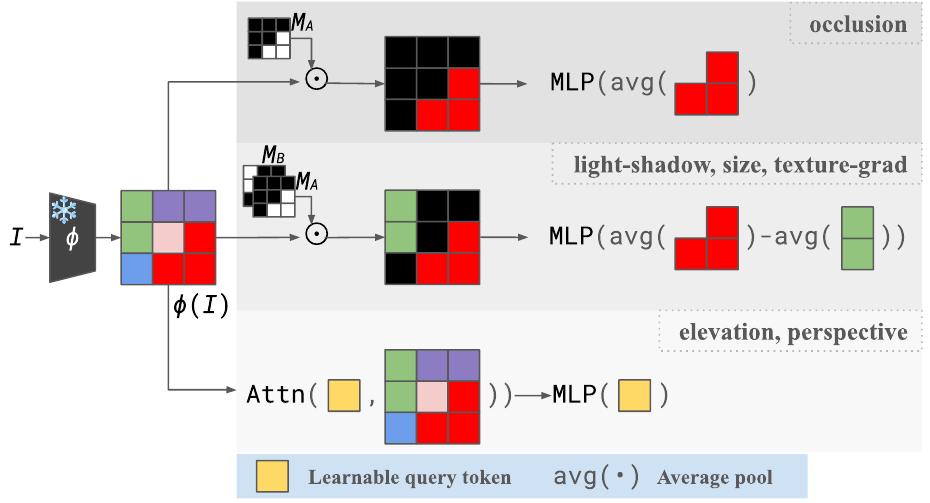}
    \vspace{-17pt}
    \caption{
        \textbf{Method for probing vision models on \name.} We extract task-specific features from the image $I$ using the vision model $\phi(\odot)$ and object masks where applicable, then train the MLP or attentive probe to solve the tasks in \name. 
        For illustrative purposes, here we only show $3\times3$ image features, but in practice, the spatial resolution is much higher. 
    }
    \vspace{-8pt}
    \label{fig:probing}
\end{figure}

\subsection{Probing Experiments} 
\label{supp:implementation_probing}

\noindent\textbf{Probing Method.} To evaluate vision models on \name, we adopt a probing approach. While the task-specific feature extraction procedure and the probe models have been discussed in the paper,  we illustrate the processes in \cref{fig:probing}. 

\noindent\textbf{Training Settings.} The MLP probes for light-shadow, occlusion, size, and texture-grad are trained with a binary cross-entropy loss since the associated tasks are binary classification ones. The attentive probes for elevation and perspective are trained with mean squared error loss due to their regression nature. All the MLP probes (for all tasks and all models) are trained for approximately 30k iterations with a batch size of eight, and the attentive probes are trained for 3,750 iterations with a batch size of 64. We used the AdamW~\cite{kingma2015adam} optimizer with cosine learning rate decay. For the hyperparameter search over model layers, we follow~\cite{el2024probing} and partition the networks into four equal-sized chunks, and evaluate the features from each chunk. For example, for DINOv2-b14, we search over layers 3, 6, 9, and 12. 

\noindent\textbf{Computational Cost.} We provide an estimation of the computational cost for evaluating a model based on the ViT-Base architecture on \name, using the standard implementation\footnote{\url{https://github.com/facebookresearch/dino}.} of DINO~\cite{caron2021emerging}. Based on the statistics of our experiment runs, on a single NVIDIA RTX A5000 GPU (24GB VRAM), benchmarking a ViT-Base on \name under our protocol and settings takes approximately 241 hours. This includes 92.7 hours for searching for the best layer for each cue (24 training + validation runs: 6 cues $\times$ 4 layers), and 148.3 hours for repeating the probing of the best layer (30 training + test runs: 6 cues $\times$ 5 repeats for statistical robustness). 
Note, these timings assume that the selected layer is the last one (\ie layer 12 for a ViT-Base model), and the total time can be lower than the estimate if this is not the case and the model is truncated up to the desired layer.

\begin{figure}[t]
    \includegraphics[width=1.0\linewidth]{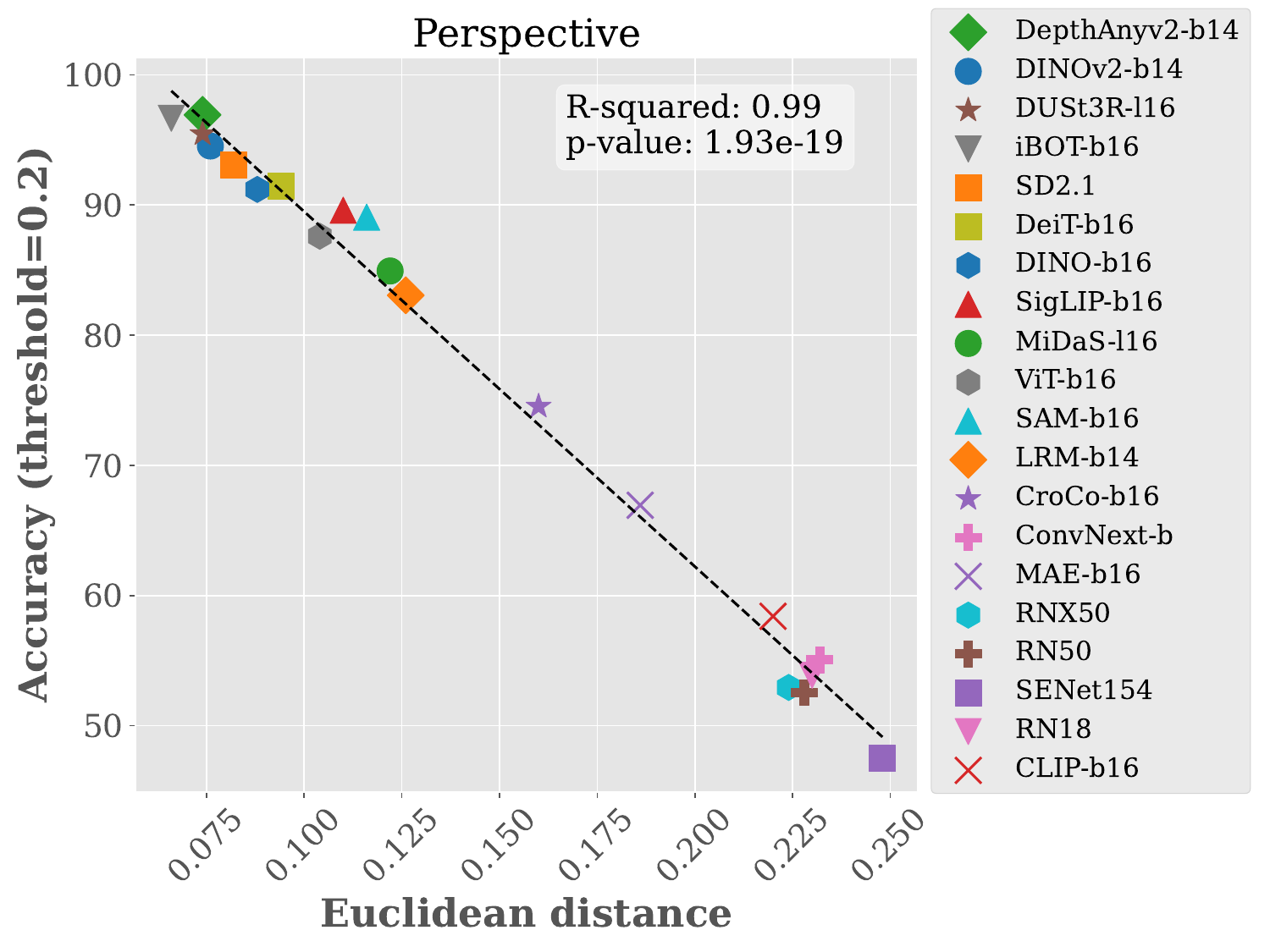}
    \vspace{-17pt}
    \caption{
        \textbf{Validating the threshold for evaluation of perspective.} We used a threshold of 0.2 to convert the Euclidean distance between ground-truth and predicted vanishing points to accuracy.
    }
    \label{fig:threshold_perspective}
\end{figure}

\begin{figure}[t]
    \includegraphics[width=1.0\linewidth]{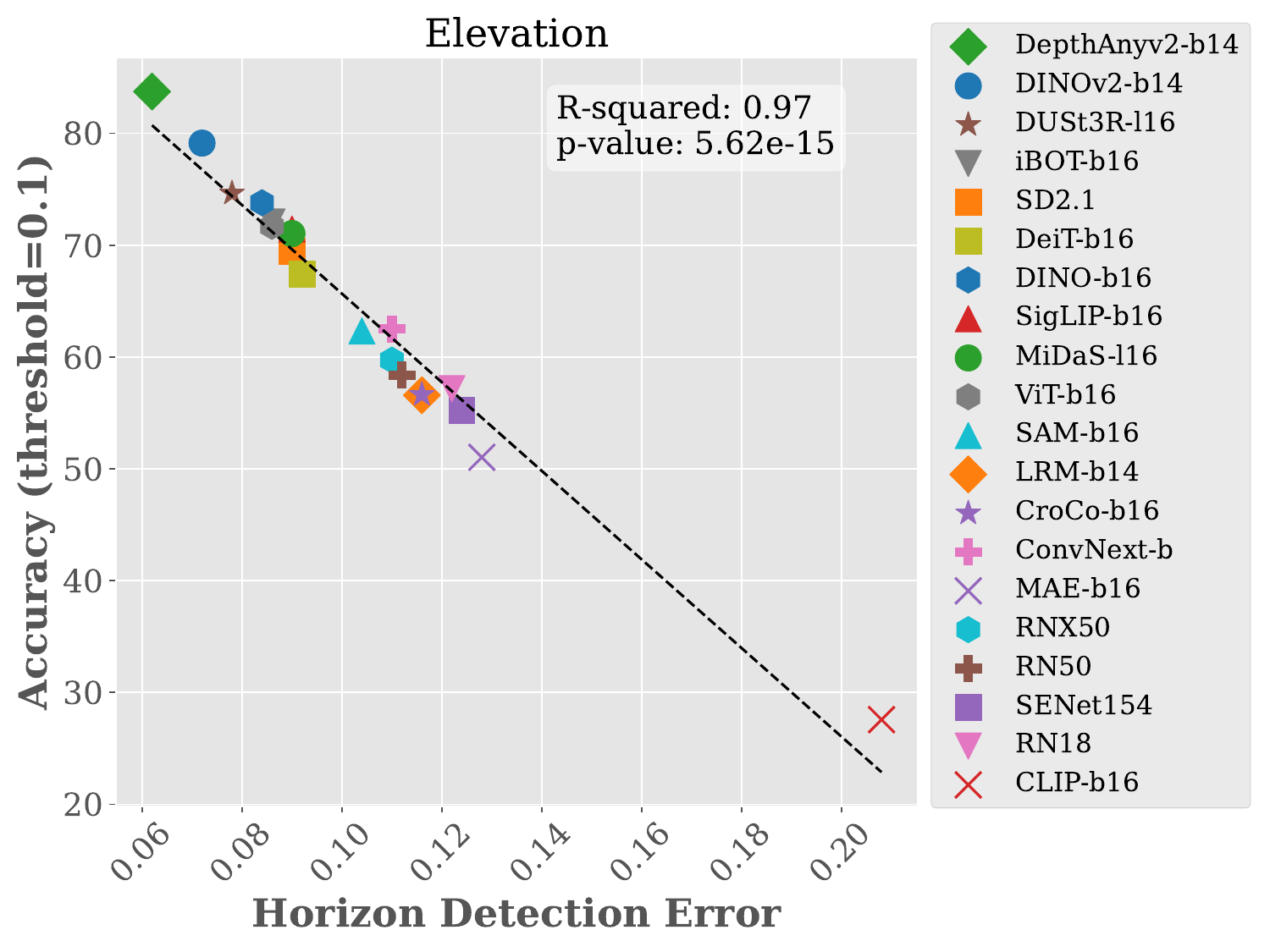}
    \vspace{-17pt}
    \caption{
        \textbf{Validating the threshold  for evaluation of elevation.} We used a threshold of 0.1 to convert the horizon detection error to accuracy.
    }
    \label{fig:threshold_elevation}
\end{figure}

\subsection{Evaluation Metrics} \label{supp:implementation_metrics}
While we use accuracy to evaluate performance on the binary classification tasks, as discussed in the main paper we apply thresholding to the Euclidean distance (for perspective) and horizon detection error\footnote{This measures the maximum distance between the predicted and the ground-truth horizon lines, normalized by the image height.}~\cite{workman2016hlw} (for elevation) to convert these metrics into accuracy. 
Specifically, we used a threshold of 0.2 for perspective and 0.1 for elevation. We validate these choices of thresholds by evaluating the correlation between the original metrics and converted accuracies in \cref{fig:threshold_perspective,fig:threshold_elevation}.

For evaluation of depth estimation on NYUv2, as in~\cite{bhat2021adabins,el2024probing}, we use accuracy, which is calculated per image using the ground-truth and predicted depth maps, $d,\hat{d}$ , as:
\begin{equation}
    \mathrm{Acc} = \frac{1}{N} \sum_{i=1}^N 
    \mathrm{max}(\frac{\hat{d}_i}{d_i},\frac{d_i}{\hat{d}_i}) < 1.25,
\end{equation}
where $N$ denotes the number of pixels in the image. For evaluation on DIW~\cite{chen2016single}, we report the Weighted Human Disagreement Rate (WHDR), which measures the proportion of depth order predictions that disagree with the manual  labels provided in DIW.

\subsection{Fine-Tuning on \textbf{\name}} \label{supp:implementation_finetune}
To fine-tune the vision models on \name we use Low Rank Adaptation (LoRA)~\cite{hu2022lora}. Here, we specify our implementation for the fine-tuned models for DINOv2. We add LoRA to the query and value projection matrices of the self-attention block in the last layer. That is, the new query $\mathbf{q}$ and value $\mathbf{v}$ for a token $\mathbf{x}$ are obtained as
\begin{gather}
   \mathbf{q} = W^Q \mathbf{x} + \Delta W^Q \mathbf{x}, \mathbf{v} = W^V \mathbf{x} + \Delta W^V \mathbf{x}, \\
   \Delta W^Q = B^Q A^Q \quad\quad \Delta W^V = B^V A^V,
\end{gather}
where $B^Q, B^V \in \mathbb{R}^{D\times R}$ and $A^Q, A^V \in \mathbb{R}^{R\times D}$. To make $\Delta W^Q, \Delta W^V$ the low-rank approximations of the original query and value projection matrices, we set $R$ to 4, which is much smaller compared to the original feature dimension $D$ (768 for DINOv2) of the model.

To train the models on \name tasks, we attach a two-layer MLP with an intermediate GELU activation~\cite{hendrycks2016gaussian} and output neurons corresponding to the tasks. The models (LoRA weights and the MLP) are trained for around 15,000 iterations using AdamW~\cite{kingma2015adam} optimizer with cosine learning rate decay and a batch size of 32. The learning rate is set to $10^{-5}$ at the start and warms up to $10^{-3}$.

\section{Evaluated Models}
\label{supp:selected_models}

\begin{table*}[t]
\centering
\resizebox{1.0\linewidth}{!}{
\begin{tabular}{l l l l r}
\toprule
\textbf{m-Rank} & \textbf{Model} & \textbf{Architecture} & \textbf{Supervision} & \textbf{Dataset (Size)}\\
\midrule
17.5          & ResNet18~\cite{he2016deep}          & ResNet      & Category    & ImageNet (1.2M)  \\
16          & ResNet50~\cite{he2016deep}         & ResNet      & Category    & ImageNet (1.2M)  \\
15          & ResNext50~\cite{xie2017aggregated}         & ResNet      & Category    & ImageNet (1.2M) \\
16.5        & SENet~\cite{hu2018squeeze}         & ResNet      & Category    & ImageNet (1.2M)    \\
11.5          & ViT~\cite{dosovitskiy2020vit} & ViT-B/16      & Category    & ImageNet (14M)    \\
6.5          & DeiT III~\cite{touvron2022deit}         & ViT-B/16      & Category    & ImageNet (14M)   \\
12.5          & ConvNext~\cite{liu2022convnet}          & CNXT-B/16    & Category             & ImageNet (14M)   \\
17.5          & MAE~\cite{he2022masked}                 & ViT-B/16      & Self-Supervised              & ImageNet (1.2M)  \\
5          & iBOT~\cite{zhou2021ibot}                & ViT-B/16      & Self-Supervised               & ImageNet (14M)  \\
9          & DINO~\cite{caron2021emerging}           & ViT-B/16      & Self-Supervised               & ImageNet (1.2M) \\
2          & DINOv2~\cite{oquab2024dinov}          & ViT-B/14      & Self-Supervised               & LVD (142M) \\
6.5          & StableDiffusion~\cite{rombach2022high}  & UNet          & Language       & LAION (5B)  \\
8          & MiDaS~\cite{ranftl2020towards}          & ViT-L/16      & Depth             & MIX-6  (1.9M)       \\
1          & DepthAnythingv2~\cite{depth_anything_v2}          & ViT-B/14      & Depth             &  MIX-13 (0.5M+62M)         \\
14.5          & LRM~\cite{hong2024lrm}                  & ViT-B/14             &  Multi-View \& 3D       & Objaverse (10M), MVImgNet (6.5M)        \\
14          & CroCo~\cite{weinzaepfel2022croco}       & ViT-B/16            & Multi-View \& Self-Supervised               & Habitat (1.8M)         \\
3.5          & DUSt3R~\cite{wang2024dust3r}          & ViT-L/16      & Multi-View \& 3D             & MIX-8 (8.5M)        \\
11.5          & SAM~\cite{kirillov2023segment}              & ViT-B/16      & Segmentation      & SA (1B)         \\
20          & CLIP~\cite{radford2021learning}         & ViT-B/16      & Language               & LAION (2B)      \\
7.5          & SigLIP~\cite{zhai2023sigmoid}           & ViT-B/16      & Language               & WebLI (18B)        \\
\bottomrule
\end{tabular}
}
\vspace{-5pt}
\caption{
    \textbf{Evaluated vision models.} We consider a range of publicly available large vision models that span several forms of supervision. In most cases, we select checkpoints of comparable model and training size. We also report the median rank (m-Rank) of each model, which is calculated based on the model's rank for each cue in \name.
}
\label{tab:models_used}
\end{table*}

Here we describe each model we evaluated in \name, grouped by the supervision type. We summarized the architecture, supervision, and training dataset, along with the median rank (m-Rank) of each model in \cref{tab:models_used}. The median rank is calculated based on the model's rank for each cue. For all models, we used publicly available checkpoints provided by the official codebases or from the timm~\cite{rw2019timm} or transformers~\cite{wolf-etal-2020-transformers} Python libraries.

\noindent\textbf{Categorization.} The following models are trained for image classification problems with category labels using the ImageNet dataset~\cite{deng2009imagenet}. ResNet18 (RN18) and ResNet50 (RN50)~\cite{he2016deep} are convolutional neural networks with residual connections. ResNext50 (RNX50)~\cite{xie2017aggregated} and SENet (SENet154)~\cite{hu2018squeeze} are extensions of residual convolutional networks, offering improved efficiency and scalability. ConvNext (ConvNext-b)~\cite{liu2022convnet} is also designed with convolutional blocks, but its design choices are reconsidered based on the success of recent transformer-based~\cite{dosovitskiy2020vit} models. ViT (ViT-b16)~\cite{dosovitskiy2020vit} and DeiT III (DeiT-b16)~\cite{touvron2022deit} are transformer-based image models built on multi-head attention. We use the base configuration with 16 patch sizes for both models.

\noindent\textbf{Depth.} MiDaS (MiDaS-l16)~\cite{ranftl2020towards} and DepthAnythingv2 (DepthAnyv2-b14)~\cite{depth_anything_v2} are models trained with dense depth supervision. Both models use transformer architectures and are trained on a mix of datasets containing dense depth supervision for pixel values. The encoder network of DepthAnythingv2 is initialized with DINOv2~\cite{oquab2024dinov}, while MiDaS is trained from scratch.

\noindent\textbf{Segmentation.} SAM (SAM-b16)~\cite{kirillov2023segment} is trained on a large-scale segmentation dataset that provides dense pixel-level category information.

\noindent\textbf{Language.}  CLIP (CLIP-b16)~\cite{radford2021learning} and SigLIP (SigLIP-b16)~\cite{zhai2023sigmoid} are trained to align the representations of images with their textual descriptions with a contrastive objective. StableDiffusion (SD2.1)~\cite{rombach2022high} is a diffusion-based generative network that produces images conditioned on text descriptions. It is trained on a large-scale dataset\cite{schuhmann2022laion} containing image-text pairs.

\noindent\textbf{Multi-View.} CroCo (CroCo-b16)~\cite{weinzaepfel2022croco} is trained with a cross-view completion objective using multi-view images, where the task involves predicting a patch of an image from another view. DUSt3R (DUSt3R-l16)~\cite{wang2024dust3r} addresses the 3D reconstruction task for the generalized stereo case using neural networks through direct regression. It takes multi-view images as input and predicts dense 2D-3D point mappings for each view. LRM (LRM-b14)~\cite{hong2024lrm} is a large-scale 3D reconstruction network that takes images as input and outputs 3D representations. The network is trained with direct supervision using a large-scale 3D object repository~\cite{deitke2024objaverse}.

\noindent\textbf{Self-Supervised.} MAE (MAE-b16)~\cite{he2022masked} and DINO (DINO-b16)~\cite{caron2021emerging} are trained with masked and contrastive self-supervised objective terms, respectively, without using any human-provided labels. iBOT (iBOT-b16)~\cite{zhou2021ibot} and DINOv2 (DINOv2-b14)~\cite{oquab2024dinov} combine masked and contrastive objectives to train networks. All of these methods are based on the transformer~\cite{dosovitskiy2020vit} architecture with the base configuration.

\clearpage